\documentclass[10pt,twocolumn,letterpaper]{article}

\usepackage{iccv}              

\definecolor{iccvblue}{rgb}{0.21,0.49,0.74}
\usepackage[pagebackref,breaklinks,colorlinks,allcolors=iccvblue]{hyperref}
\usepackage{multicol}
\usepackage{multirow}
\usepackage{amssymb}
\usepackage{pifont}
\usepackage{longtable}
\usepackage{tcolorbox}

\definecolor{Highlight}{HTML}{39b54a}
\newcommand{\hl}[1]{\textcolor{Highlight}{#1}}
\newcommand{\hlr}[1]{\textcolor{red}{#1}}
\usepackage{graphicx}
\usepackage{amsmath, graphicx, amssymb}
\usepackage{csquotes}
\usepackage{colortbl}

\usepackage{tabularx}      
\usepackage{booktabs}      
\usepackage{ragged2e}      
\usepackage{array}         

\definecolor{color_1}{RGB}{178,86,40}    
\definecolor{color_2}{RGB}{55,83,156}      
\definecolor{color_3}{RGB}{38,79,34} 

\def\paperID{10958} 
\def\confName{ICCV}
\def\confYear{2025}

\title{
\raisebox{-.25\height}{
\includegraphics[width=0.045\textwidth]{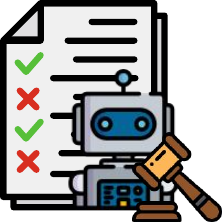}
}ProJudge: A Multi-Modal Multi-Discipline Benchmark and Instruction-Tuning Dataset for MLLM-based Process Judges}

\author{
  Jiaxin Ai\textsuperscript{1,2},
  Pengfei Zhou\textsuperscript{3},
  Zhaopan Xu\textsuperscript{3},
  Ming Li\textsuperscript{3},
  Fanrui Zhang\textsuperscript{4,2},
  Zizhen Li\textsuperscript{5,2},
  Jianwen Sun\textsuperscript{5,2},\\
  Yukang Feng\textsuperscript{5,2},
  Baojin Huang\textsuperscript{6},
  Zhongyuan Wang\textsuperscript{$\dagger$}\textsuperscript{1},
  Kaipeng Zhang\textsuperscript{$\dagger$}\textsuperscript{3,2} \\
  \textsuperscript{1}WHU, 
  \textsuperscript{2}Shanghai Innovation Institude,
  \textsuperscript{3}Shanghai AI Laboratory,
  \textsuperscript{4}USTC,
  \textsuperscript{5}NKU,
  \textsuperscript{6}HZAU\\
  \tt\small
  julyai@whu.edu.cn, 
  zhangkaipeng@pjlab.org.cn\textsuperscript{$\dagger$}, \\
  \tt\small
  \url{https://github.com/jiaxin-ai/ProJudge}
}

\begin{document}
\maketitle
\begin{abstract}
As multi-modal large language models~(MLLMs) 
frequently exhibit errors when solving scientific problems, evaluating the validity of their reasoning processes is critical for ensuring reliability and uncovering fine-grained model weaknesses. Since human evaluation is laborious and costly, prompting MLLMs as automated process judges has become a common practice.
However, the reliability of these model-based judges remains uncertain.
To address this, we introduce \textbf{ProJudgeBench}, the first comprehensive benchmark specifically designed for evaluating abilities of MLLM-based process judges.
ProJudgeBench comprises 2,400 test cases and 50,118 step-level labels, spanning four scientific disciplines with diverse difficulty levels and multi-modal content. 
In ProJudgeBench, each step is meticulously annotated by human experts for correctness, error type, and explanation, enabling a systematic evaluation of judges' capabilities to detect, classify and diagnose errors. 
Evaluation on ProJudgeBench reveals a significant performance gap between open-source and proprietary models. 
To bridge this gap, we further propose \textbf{ProJudge-173k}, a large-scale instruction-tuning dataset, and a \textbf{Dynamic Dual-Phase} fine-tuning strategy that encourages models to explicitly reason through problem-solving before assessing solutions. Both contributions significantly enhance the process evaluation capabilities of open-source models.
All the resources will be released to foster future research of reliable multi-modal process evaluation.
\end{abstract}

\vspace{-2mm}  
\section{Introduction}
\begin{figure}[!t]
\centering
  \resizebox{1\linewidth}{!} {
    \includegraphics{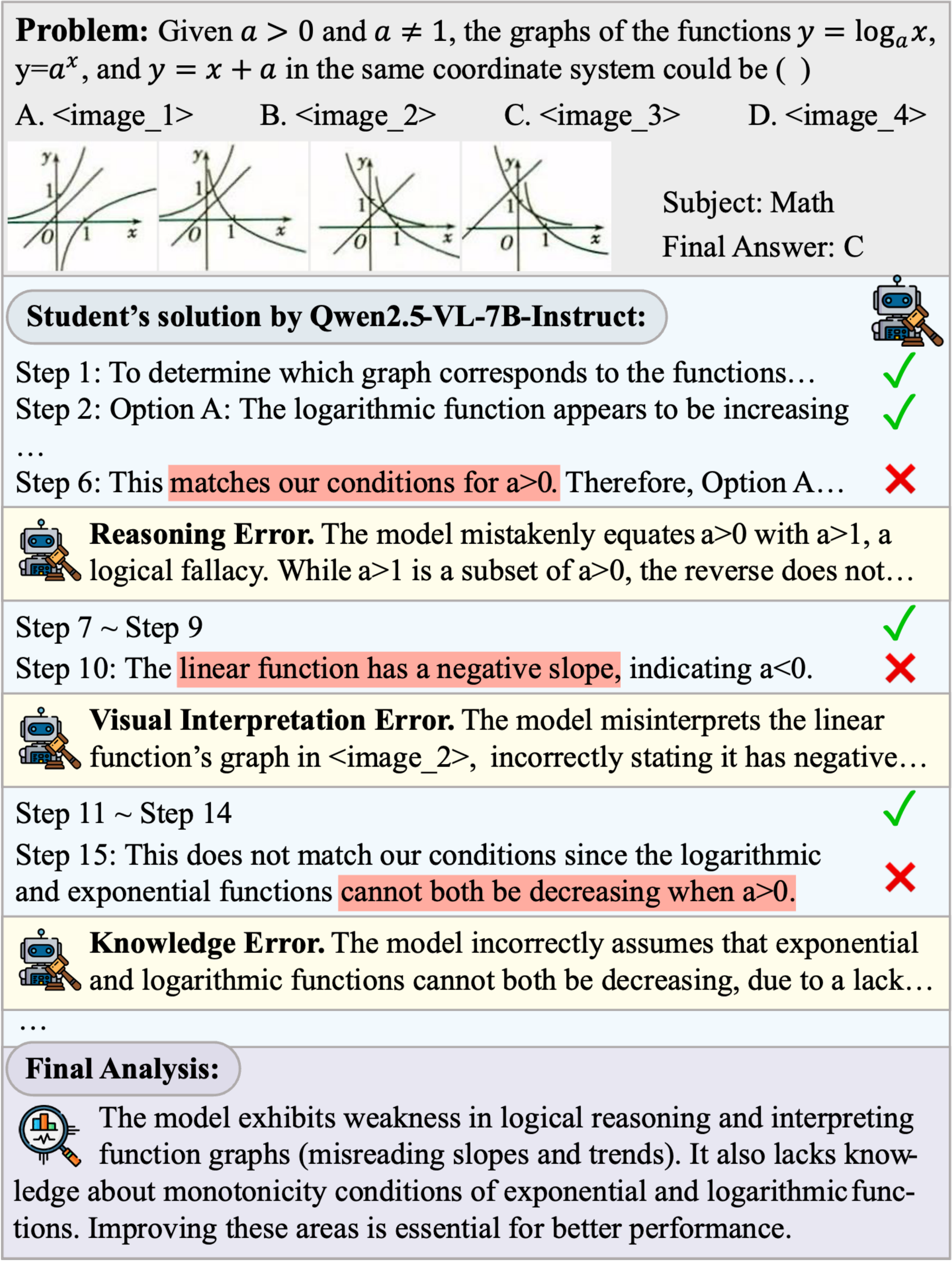}
  }
  \vspace{-5mm}
  \caption{Task definition of process evaluation. For each step, MLLM-based process judges detect errors, classify error types and provide brief explanations. 
  Based on these analyses, we derive insights into model weaknesses, guiding future improvements.
  }
    \label{fig:task_definition}
  \vspace{-5mm}
\end{figure}

Recent multi-modal large language models (MLLMs)~\cite{ref:gpt4o,ref:o1,ref:gemini_2_flash,ref:qvq} have demonstrated remarkable capabilities in solving scientific problems.
While final answer accuracy is commonly used to evaluate performance~\cite{ref:mmmu,ref:gsm8k,ref:math,ref:olympicarena}, the validity of the solving process itself is equally critical~\cite{ref:reasoneval,ref:uesato2022solving,ref:prm800k}. 
Without examining intermediate steps, models may produce correct answers through flawed reasoning, leading to an overestimation of their true capabilities. 
Besides, compared with final answer evaluation, fine-grained error analyses of reasoning process can reveal more detailed model weaknesses, providing actionable insights for targeted improvements.
These factors highlight the necessity of \textbf{Process Evaluation}.

\newcommand{\tabincell}[2]{\begin{tabular}{@{}#1@{}}#2\end{tabular}}
\begin{table*}[t]
  \centering
  \vspace{1mm}
  \resizebox{\textwidth}{!}{
    \begin{tabular}{lcccccccc}
        \toprule
         & \tabincell{c}{\textbf{Process Judge}\\ \textbf{Benchmark} } 
         & \tabincell{c}{\textbf{Multi-modal}\\ \textbf{Benchmark}}
         & \tabincell{c}{\textbf{Multi-Discipline} \\ \textbf{Benchmark}} 
         & \tabincell{c}{\textbf{Multi-Difficulty} \\ \textbf{Problems}}   
         & \tabincell{c}{\textbf{Step-level} \\ \textbf{Annotation}}   
         & \tabincell{c}{\textbf{Fine-grained} \\ \textbf{Error Analysis}}   
         & \tabincell{c}{\textbf{Avg.} \\ \textbf{Steps}}   \\
        \midrule
        MR-GSM8K \cite{ref:mr_gsm8k}& \ding{55} & \ding{55} & \ding{55} & \ding{55} & \checkmark & \ding{55} & 8.3\\
        CriticBench \cite{ref:criticbench}& \ding{55} & \ding{55} & \checkmark & \ding{55} & \ding{55} & \ding{55} & -\\
        MathCheck-GSM \cite{ref:mathcheck_gsm}& \ding{55} & \ding{55} & \ding{55} & \ding{55} & \checkmark & \ding{55} & -\\
        ProcessBench \cite{ref:processbench}& \ding{55} & \ding{55} & \ding{55} & \checkmark & \checkmark & \ding{55} & 7.1 \\
        PRMBench~\cite{ref:prmbench}& \ding{55} & \ding{55} & \ding{55} & \ding{55} & \checkmark & \checkmark & 13.4 \\
        \midrule
        \textsc{ProJudgeBench}& \checkmark & \checkmark & \checkmark & \checkmark & \checkmark & \checkmark & 20.8  \\ 
        \bottomrule
    \end{tabular}
  }
  \vspace{-1mm}
  \caption{Comparison between related benchmarks with our ProJudgeBench.}
  \label{tab:comparison}
  \vspace{-4.5mm}
\end{table*}

As MLLMs become more advanced, evaluating lengthy and intricate reasoning processes becomes increasingly challenging for humans. 
As a result, researchers are turning to prompt MLLMs as automated process judges~\cite{ref:mm-math,ref:olympicarena,ref:mathverse,ref:mme-cot}.
While promising, this also raises a critical question: \textit{How reliable are these model-based judges?} 
Since MLLMs may be prone to errors and biases, their abilities to conduct accurate and objective process evaluation remains uncertain. 

However, there is currently a lack of comprehensive benchmarks specifically designed to assess the capabilities of these model-based process judges. Although some existing benchmarks~\cite{ref:mr_gsm8k,ref:criticbench,ref:mathcheck_gsm,ref:processbench,ref:prmbench} can be adapted for this purpose, they suffer from three critical limitations:
(1) \textbf{Narrow scope}: Most are limited to a single modalitiy, single discipline, or problems of limited difficulty, failing to capture the diverse challenges of real-world reasoning tasks.
(2) \textbf{Insufficient error analysis}: They primarily focus on error detection, while neglecting the assessing of judges' nuanced error diagnosing capabilities—vital for enhancing judging explainability and uncovering model weaknesses.
(3) \textbf{Synthetic or Non-Generalizable Data}: Many benchmarks are either tailored to specific policy models, or based on synthetic error data (e.g. GPT-modified errors) that fail to reflect diverse error patterns observed in actual model solutions, limiting their applicability in real-world scenarios.

To address these limitations, we introduce \textbf{ProJudgeBench}, a comprehensive benchmark specifically designed for assessing abilities of MLLM-based process judges.
ProJudgeBench possesses the following distinctive features:
(1) \textbf{Multi-modal, Multi-discipline, and Multi-difficulty}: It comprises 2,400 test cases and 50,118 step-level labels, spanning mathematics, physics, chemistry, and biology, with diverse difficulty levels and multi-modal content.
(2) \textbf{Fine-grained Error Analysis}: We define seven error types that encompass common mistakes models may make in long reasoning chains. Each step is meticulously annotated by human experts for correctness, error type, and explanation, enabling a systematic evaluation of process judges' capabilities to detect, classify, and diagnose errors in scientific problem-solving tasks.
(3) \textbf{Realistic and Diverse Error Patterns}: 
To ensure real-world applicability, we collect solutions from 10 distinct MLLMs of varying sizes, architectures, and design goals, reflecting a broad spectrum of realistic reasoning behaviors and error patterns.

We evaluate 11 MLLMs on ProJudgeBench, revealing a significant performance gap between proprietary and open-source models. 
To bridge this gap, we further present \textbf{ProJudge-173k}, 
a large-scale instruction tuning dataset designed for fine-grained evaluation of step-by-step reasoning. 
The dataset is constructed via two complementary pathways, 
ensuring both diversity and real-world relevance. 
Rigorous filtering guarantees high-quality synthetic annotations, providing a foundation for fine-tuning open-source models as process judges.
Besides, we propose a \textbf{Dynamic Dual-Phase (DDP)} fine-tuning strategy that encourages models to explicitly reason through problem-solving steps before assessing solutions, mimicking the behavior of human experts. Such explicit thinking not only deepens the model's understanding of the problems but also improves the robustness and generalizability of process judges.
By fine-tuning on ProJudge-173k with the proposed  DDP strategy, we demonstrate significant improvements in process evaluation capabilities of open-source models, narrowing their performance disparity with proprietary systems.

Our contributions can be summarized as follows:
\begin{itemize}
    \item We introduce \textbf{ProJudgeBench}, a multi-modal, multi-discipline benchmark specifically designed for assessing fine-grained error detection, classification and diagnosis capabilities of MLLM-based process judges.
    \item To bridge the gap in process evaluation capabilities between open-source and proprietary models, we further present \textbf{ProJudge-173k}, a large-scale instruction tuning dataset, and a \textbf{Dynamic Dual-Phase} fine-tuning strategy. Both of these innovations significantly improve the process evaluation capabilities of open-source models.
    \item Comprehensive experiments uncover key challenges and limitations in current models, providing valuable insights into multi-modal reasoning and process evaluation.
\end{itemize}

\begin{figure*}[!t]
\centering
  \resizebox{0.97\linewidth}{!} {
    \includegraphics{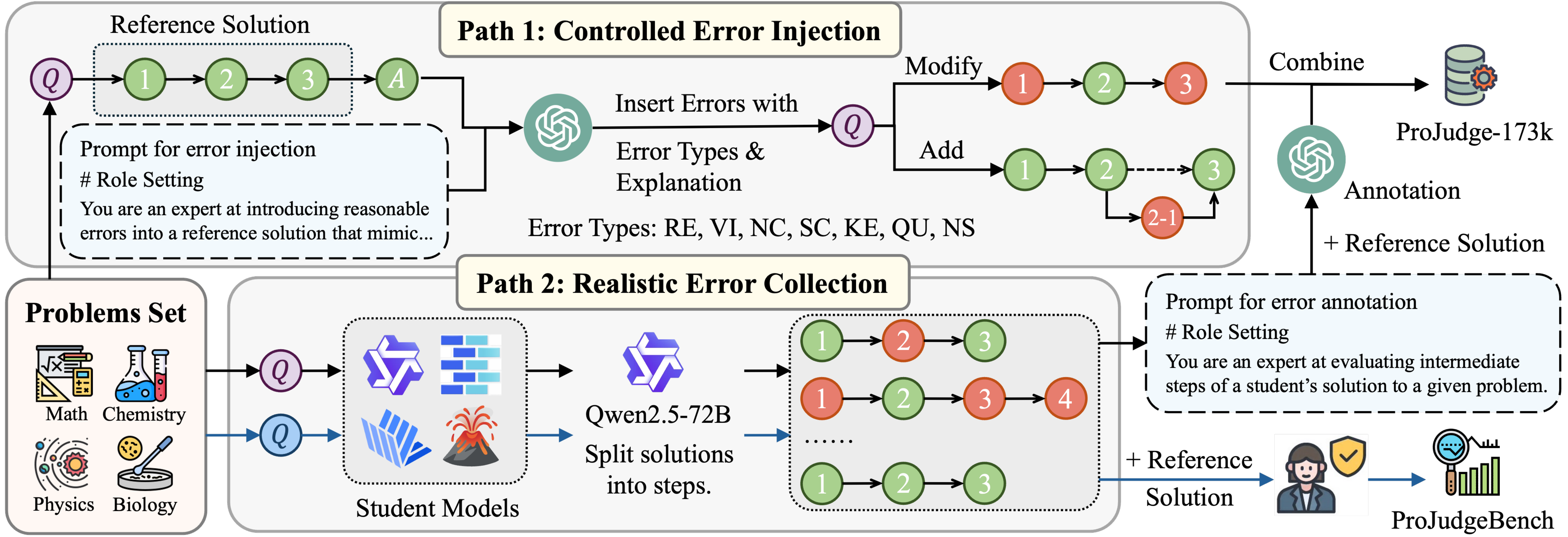}
  }
  \vspace{-2mm}
  \caption{An overview for data construcion process of ProJudgeBench and ProJudge-173k.
  }
    \label{fig:main}
  \vspace{-4.5mm}
\end{figure*}
\section{Related Work}
\noindent \textbf{Multi-Modal Benchmarks.}
With the fast development of recent MLLMs~\cite{ref:gemini_2_flash,ref:gpt4o,ref:qwen2.5-vl,ref:blip,ref:deepseekvl}, it becomes essential to evaluate their performance across a variety of tasks to understand their strengths and weaknesses~\cite{ref:mmmu,ref:mathvista,ref:liu2024mm,ref:eee-bench,ref:lvlm,ref:mmbench,ref:mmt-bench,masry2022chartqa,ref:seed-bench,ref:ok-bench,ref:docvqa,ref:wemath,qiao2024we,ref:math-vision,wang2025charxiv}.
For example, MMMU~\cite{ref:mmmu} has been proposed to evaluate the scientific problem-solving abilities of MLLMs across a range of college-level subjects. MathVista~\cite{ref:mathvista} is widely used to assess the mathematical capabilities of MLLMs. While considerable progress has been achieved, these benchmarks often focus solely on evaluating final answers, overlooking crucial information of intermediate reasoning steps and potentially leading to unreliable results.

\noindent \textbf{MLLM-based Process Judges.}
MLLM-based process judges have been widely used to automatically evaluate multi-modal reasoning steps of large language models~\cite{ref:mm-math,ref:olympicarena,ref:mathverse,ref:mme-cot}.
For example, 
MM-Math~\cite{ref:mm-math} incorporates MLLM-as-a-judge to automatically analyze intermediate steps and identify errors.
OlympicArena~\cite{ref:olympicarena} employs GPT-4V~\cite{ref:gpt4v} to score the correctness of each solution step, ensuring a rigorous assessment.
MathVerse~\cite{ref:mathverse} utilizes GPT-4V~\cite{ref:gpt4v} to extract and assess key reasoning steps, providing nuanced scoring.
Despite the widespread adoption, the reliability of MLLM-based judges is rarely scrutinized. 
Besides, current evaluation predominantly relies on proprietary models, suffering from prohibitive costs and reproducibility instability. This necessitates the development of open-source process judges. 
To address these challenges, we introduce ProJudgeBench, a comprehensive benchmark for assessing MLLMs' capabilities as process judges, and ProJudge-173k, a large-scale instruction-tuning dataset designed to enhance open-source MLLMs' process evaluation abilities.

\noindent \textbf{Process Evaluation Benchmarks.}
There exist several benchmarks related to assessing process evaluation abilities of LLMs~\cite{ref:mr_gsm8k,ref:mathcheck_gsm,ref:criticbench,ref:processbench,ref:prmbench}.
For example, ProcessBench~\cite{ref:processbench} measures the ability to identify erroneous steps in mathematical reasoning.
PRMBench~\cite{ref:prmbench} synthesize erroneous steps based on PRM800k~\cite{ref:prm800k}, evaluating fine-grained error detection capabilities of PRMs across multiple dimensions.
Our ProJudgeBench is distinguished from prior benchmarks in three key aspects, as shown in Table~\ref{tab:comparison}. 
First, it covers four scientific discipline with multi-modal content and varying difficutly levels, reflecting the complexity of real-world reasoning tasks.
Second, test cases are curated from diverse model-generated solutions instead of synthetic data, capturing a wide range of realistic reasoning behaviors and error patterns.
Third, each step is human-annotated for correctness, error type and explanation, enabling fine-grained evaluation of models' error-diagnosing capabilities.
\section{ProJudgeBench}
\subsection{Task Definition}
\label{sec:task_definition}
As shown in Figure~\ref{fig:task_definition}, given a scientific problem \(P\) with its final answer \(A\), and a step-by-step solution \(S=\{s_0, s_1, \cdots, s_{n-1}\}\) generated by a student model, ProJudgeBench requires judge models to perform fine-grained evaluations of each step \(s_i\) by determining its correctness, classifying the error type and providing a brief explanation. Specifically, for each step \(s_i\), the model outputs a tuple \((c_i, e_i, r_i)\), where \(c_i \in \{1, 0\}\) indicates whether the step is correct (\(c_i=1\)) or not (\(c_i=0\)); \(e_i\) denotes the error type; and \(r_i\) is the natural language explanation of the error cause.

To categorize errors, we define 7 error types based on a thorough analysis of common mistakes that models tend to make during long-chain reasoning processes: \textbf{reasoning error}, \textbf{visual interpretation error}, \textbf{numerical calculation error}, \textbf{symbolic calculation error}, \textbf{knowledge error}, \textbf{question understanding error}, and \textbf{no solution provided}. Definitions of each category are displayed in Appendix~\ref{appendix_error_types_definition}.

Unlike existing benchmarks that focus solely on identifying the initial error location~\cite{ref:mm-math,ref:processbench,ref:prm800k,ref:math_shepherd}, ProJudgeBench emphasizes a fine-grained analysis of the entire reasoning process. This is motivated by the observation that, model-generated solutions often contain multiple errors of diverse types distributed across different steps. By requiring models to not only detect errors but also classify their types and diagnose their root causes, ProJudgeBench can provide a more comprehensive judging tool for uncovering the weaknesses of student models and thereby providing actionable insights to guide their improvement.

\subsection{Data Construction}
\label{sec:ProJudgeBench_data_construction}
\noindent \textbf{Problem Curation.}
We collect problems from 
two public benchmarks, OlympiadBench~\cite{ref:olympiadbench} and OlympicArena~\cite{ref:olympicarena}, covering four scientific disciplines
and various modalities.
Since both benchmarks focus on college and competition-level problems, 
we supplement the dataset with problems from primary, middle, and high school (referred to as K12) to ensure more diverse difficulty levels.
In total, we curate 400 scientific problems, with a balanced distribution across disciplines and difficulty levels, providing a comprehensive set for analyzing scientific problem-solving processes.

\noindent \textbf{Step-by-Step Solution Generation.}
We generate solutions using various proprietary and open-source MLLMs, including GPT-4o~\cite{ref:gpt4o}, InternVL series~\cite{ref:internvl2.5}, QwenVL series~\cite{ref:qwen2.5-vl}, LLaVA series~\cite{ref:llava} and MiniCPM-V~\cite{ref:minicpm}, resulting in a total of 10 distinct solution generators. 
Each problem is solved by 5 models randomly selected from the 10 candidates, with strict balancing mechanism to ensure equal usage across all models.
The complete list of models and generating prompts
are provided in Appendix~\ref{appendix_data_construction_ProJudgeBench}. 
To standardize the step granularity of each solution, we prompt Qwen2.5-72B-Instruct~\cite{ref:qwen2.5} to split solutions into 
steps. This reformatting process ensures consistency in step segmentation, which is crucial for accurate human annotation. Solutions that were altered during reformatting (e.g., changes in final answers) are manually segmented to maintain the integrity.
Additionally, we collect ground-truth solutions for each problem, where every step is labeled with \(c_i=1\).

\noindent \textbf{Expert Annotation.}
We recruit 6 human experts with domain expertise (at the undergraduate level or above) for annotation. All of them are required to pass a mandatory proficiency examination and complete an annotation tutorial. The annotators are tasked with evaluating each step in the model-generated solutions for correctness, error type, and root cause.
To reduce annotation difficulty and ensure accuracy, we provide annotators with a reference solution for each problem, helping them to first understand the correct problem-solving approach before assessing the model-generated solutions.
To ensure annotation reliability, we implement a rigorous quality control mechanism. Please refer to Appendix~\ref{appendix_quality_control_ProJudgeBench} for details about quality control process.




\subsection{Statistics}
\label{sec:ProJudgeBench_statistics}
\begin{figure}[!t]
\centering
  \resizebox{1.0\linewidth}{!} {
    \includegraphics{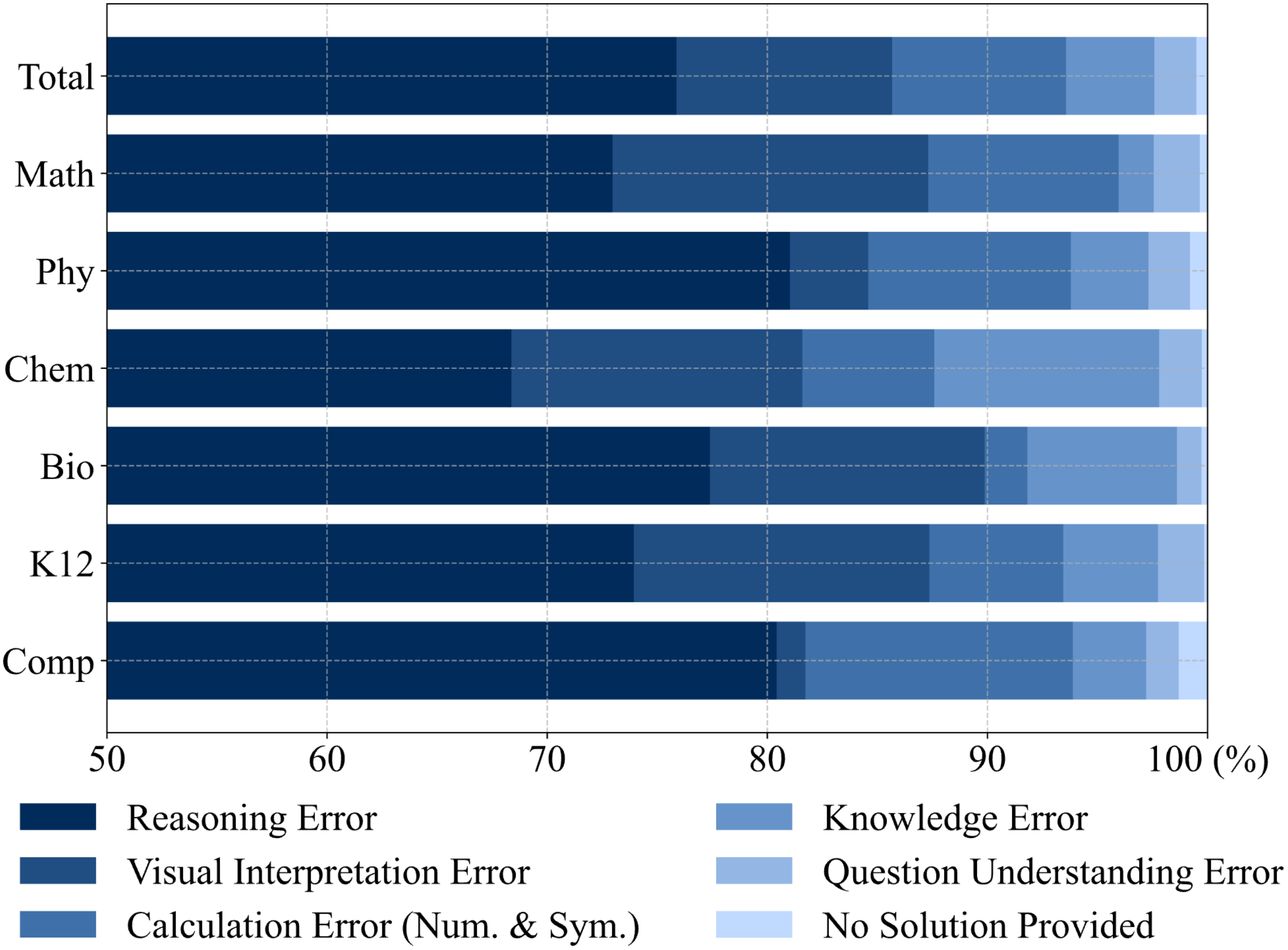}
  }
  \vspace{-5mm}
  \caption{Distribution of error types across different disciplines and difficulty levels in ProJudgeBench.
  K12 and Comp represent routine and competition-level problems, respectively.
  }
    \label{fig:error_types_distribution}
  \vspace{-4mm}
\end{figure}

The final ProJudgebench comprises 2400 test cases, with detailed statistics presented in Table~\ref{tab:statistics}.
Generally, higher-difficulty problems often lead to more steps in model-generated solutions, with increased error rates and more diverse error types.
We analyze the distribution of error types across different disciplines and difficulty levels in Figure~\ref{fig:error_types_distribution}. Reasoning errors are the most prevalent, accounting for the highest proportion of errors overall. Notably, reasoning errors are even more dominant in high-difficulty problems compared to routine problems, highlighting the challenges models face in handling complex logical reasoning. Visual interpretation and calculation errors follow in frequency, suggesting that models still struggle with accurately interpreting visual information and performing precise calculations. In biology and chemistry problems, knowledge-based errors are more prominent, reflecting the models' limitations in domain-specific understanding.

\section{ProJudge-173k for Instruction Tuning}


\subsection{Data Construction}
\label{sec:ProJudge-173k_data_construction}
As shown in Figure~\ref{fig:main}, we adopt two complementary paths to construct the dataset, 
each addressing different aspects of error generation and evaluation.

\noindent \textbf{Path 1: Controlled Error Injection.}
In this process, we utilize GPT-4o~\cite{ref:gpt4o} to intentionally inject errors into correct solutions. We first apply few-shot learning to familiarize the model with various error types. 
Subsequently, we instruct the model to randomly inject multiple errors into correct solutions, creating flawed step-by-step reasoning paths.
For this purpose, we select problems from the Camel~\cite{ref:camel} dataset, which is relatively easy for the model to understand and inject reasonable errors. 
Camel covers four disciplines: mathematics, physics, chemistry, and biology. For each discipline, we randomly select a subset of 2,000 problems with their ground-truth solutions, which are then split into reasoning steps for injection.
This process ensures high annotation accuracy since errors are deliberately injected with known types and causes.
However, it sacrifices some realism, as the artificially introduced errors may differ from those naturally occurring in the model’s reasoning process.

\noindent \textbf{Path 2: Realistic Error Collection.}
This path closely mirrors the ProJudgeBench construction process. We begin by collecting competition-level problems from the training set of OlympiadBench~\cite{ref:olympiadbench}, including 4,013 mathematics and 2,071 physics problems. 
To incorporate a broader range of difficulty levels, we compile a K12 dataset consisting of 4,000 problems each from mathematics, physics, chemistry, and biology, covering primary, middle and high school levels.
To simulate real-world evaluation scenarios, we generate solutions using 9 diverse MLLMs, which are then segmented into step-by-step reasoning paths.
GPT-4o~\cite{ref:gpt4o} is then prompted to evaluate each step for correctness, error types and explanations. 
While this approach heavily depends on the expertise of the annotation model, it ensures consistency with real-world evaluation tasks by capturing errors commonly encountered during model reasoning.


\noindent \textbf{Data Filtering.}
To ensure data quality, we implement rigorous filtering processes, including format consistency, annotation consistency, and error coverage checks. Please refer to Appendix~\ref{appendix_data_filtering_ProJudge-173k} for more details.


\begin{table}[t]
\centering
\resizebox{0.46\textwidth}{!}{
\begin{tabular}{lc}
\toprule
\textbf{Statistic} & \textbf{Number} \\
\midrule
\textbf{ProJudgeBench} & 2,400 \\ 
- \# Math / Phy. / Chem. / Bio. & 600 \\
- \# K12 / OlymBench / OlymArena & 1,350 / 250 / 625 \\
- Avg. / Max. Steps & 20.8 / 470 \\
- Avg. / Max. Error Types & 1.5 / 5 \\
- Avg. / Max. / \% Error Steps & 6.6 / 226 / 21.6 \\
\midrule
\textbf{ProJudge-173k} & 173,354 \\
- \# Math / Phy. / Chem. / Bio. & 58k / 40k / 37k / 35k  \\
- \# Camel / K12 / OlymdBench & 26k / 93k / 53k \\
- \# Avg. / Max. Steps & 18.2 / 926 \\
- Avg. / Max. Error Types & 1.4 / 5 \\
- Avg. / Max. / \% Error Steps & 5.9 / 402 / 24.7 \\
\bottomrule
\end{tabular}
}
\vspace{-1mm}
\caption{Statistics of ProJudgeBench and ProJudge-173k. }
\label{tab:statistics}
\vspace{-4mm}
\end{table}
\subsection{Statistics}
\label{sec:ProJudge-173k_statistics}
The statistics are presented in Table~\ref{tab:statistics}. 
Our dataset spans a wide range of difficulty levels and scientific domains, ensuring a balanced representation across various problem types and evaluation scenarios. On average, each solution contains 18 steps, with some solutions extending to over 900 steps, reflecting the complexity of real-world reasoning tasks. 
The dataset also ensures diversity in error types and modalities, making it a comprehensive resource for training MLLM-based process judges.
To the best of our knowledge,\textbf{ ProJudge-173k is the first large-scale instruction tuning dataset specifically designed for process evaluation with fine-grained step-level annotations.} 


\section{Dynamic Dual-Phase Fine-tuning}
To enhance the robustness of process judges, we propose a Dynamic Dual-Phase fine-tuning strategy, which consists of two training phases: Direct Evaluate and Synthesize-then-Evaluate. 

\subsection{Phase 1: Direct Evaluate}
This phase serves as the baseline task, where the model directly evaluates the student's solution \(x_S\) based on the problem \(x^j_\mathrm{P}\) and its final answer \(x^j_\mathrm{A}\).
For each step \(s_i\) in \(x^j_\mathrm{S}\), the model generates an evaluation result organized as a tuple \((s_i, c_i, e_i, r_i)\), denoting step description, correctness, error type, and root cause, respectively.

The training set for this task can be expressed as:
\(D_{\mathrm{DE}} = \{(x_\mathrm{P}^j, x_\mathrm{A}^j, y_{\mathrm{PE}}^j)\}_{j=1}^{N}\), where \(y^j_{\mathrm{PE}}\) represents the ground-truth step annotations and $N$ denotes the dataset size.
During training, the model minimizes the cross-entropy loss between its predictions and the ground-truth annotations:

\vspace{-6mm}
\begin{equation}
\small
\begin{aligned}
    \mathcal{L}_{\mathrm{DE}}(\theta, D_{\mathrm{DE}}) = - \frac{1}{N}\sum_{j=1}^N[ 
    \sum_{t=1}^{|y_{\mathrm{PE}}^j|} \log p
    (y_{\mathrm{PE},t}^j | x^j_\mathrm{P}, x^j_\mathrm{A}, x^j_\mathrm{S}, y^j_{\mathrm{PE}, <t}; \theta)
    ] 
\end{aligned}
\label{eq:loss1}
\end{equation}
where $y^j_{\mathrm{PE},t}$ denotes the t-th token in the ground-truth sequence, \(y^j_{\mathrm{PE},<t}\) represents the preceding tokens. 


\subsection{Phase 2: Synthesize-then-Evaluate}
In this phase, the model is encouraged to explicitly reason through problem-solving steps before evaluating student solutions, closely mimicking the behavior of human experts.
Formally, the model is first required to reason through the problem \(x^j_\mathrm{P}\) and synthesize a reference solution \(\hat{S}\), which is expected to reach the correct final answer \(x^j_\mathrm{A}\):
\(\hat{S}\sim p(*|x^j_\mathrm{P},x^j_\mathrm{A},I_\mathrm{solv};\theta)\). \(I_\mathrm{solv}=\)``Let's solve the problem step by step to get the correct final answer."

With \(\hat{S}\) generated, the model gains a deeper understanding of the problem, enabling more informed judgments. This also provides a reference for subsequent process evaluation. Based on \(\hat{S}\), the model conducts process evaluation on the student's solution \(x^j_\mathrm{S}\), formalized as:
\(\hat{PE}\sim p(*|\hat{S}, x^j_\mathrm{S},I_\mathrm{solv};\theta)\), where \(I_\mathrm{eval}=\)``Based on our solution, let's evaluate the student's solution step by step."


The training set for this phase is structured as: \(D_\mathrm{SE} = \{(x^j_\mathrm{P}, x^j_\mathrm{A}, I_\mathrm{solv}, y^j_\mathrm{S}),(\hat{S}, x^j_\mathrm{S}, y^j_\mathrm{PE}, I_\mathrm{eval})\}_{j=1}^N\), where \(y^j_\mathrm{S}\) and \(y^j_\mathrm{PE}\) denote the ground-truth solution and process evaluation results respectively. 
During training, the model is optimized on two separate cross-entropy losses to handle both the solution synthesis and the process evaluation tasks:
\vspace{-2mm}
{\small
\begin{align}
    \mathcal{L}_\mathrm{SE}(\theta, D_\mathrm{SE}) 
    &= -\frac{1}{N}\sum_{j=1}^N [ 
    \sum_{t=1}^{|y^j_\mathrm{S}|} \log p
    (y^j_{\mathrm{S},t} | x^j_\mathrm{P}, x^j_\mathrm{A}, I_\mathrm{solv}, y^j_{\mathrm{S},<t}; \theta) \nonumber\\
    &+\sum_{t=1}^{|y^j_\mathrm{PE}|} \log p(y^j_{\mathrm{PE},t} | \hat{S}, x^j_\mathrm{S}, I_\mathrm{eval}, y^j_{\mathrm{PE},<t}; \theta)]
\end{align}
}

\subsection{Dynamic Dual-Phase Training}
During fine-tuning, the model dynamically alternates between the two phases with a probability \(p\). This mechanism introduces variability and diversity into the training process, enhancing model's robustness and generalizability across a wide range of evaluation scenarios.
\begin{table*}
\belowrulesep=0pt
\aboverulesep=0pt
\fontsize{14}{21}
\selectfont
\setlength{\tabcolsep}{4pt}
\centering
  \resizebox{\textwidth}{!}{
    \begin{tabular}{lc|c|ccccccc}
    \toprule
    
\multirow{2}{*}{\textbf{Model Name}} & 
\multirow{2}{*}{\tabincell{c}{\textbf{Step}\\\textbf{Corr.} }} &
\multicolumn{8}{c}{\textbf{Error Types}}  \\ 
\cmidrule(){3-10} 
&& \textbf{Overall} & \textbf{RE.}  & \textbf{VI.}  & \textbf{NC.}  & \textbf{SC.}  & \textbf{KE.}  & \textbf{QU.}  & \textbf{NS.}  \\

\rowcolor{lightgray!20}
\midrule 
\multicolumn{10}{c}{\textbf{\textit{Open-source MLLMs}}} \\
\hline
\href{https://internvl.github.io/blog/2024-12-05-InternVL-2.5/}{\textcolor{color_1}{InternVL2.5-8B}} & 25.58 & 6.77 & 8.19 & 0. & 8.18 & 1.75 & 1.71 & 1.07 & 0. \\
\href{https://internvl.github.io/blog/2024-12-05-InternVL-2.5/}{\textcolor{color_1}{InternVL2.5-26B}} & 66.72 & 13.51 & 16.44 & 0.64 & 14.80 & 2.63 & 2.85 & 2.15 & \textbf{4.16} \\
\href{https://internvl.github.io/blog/2024-12-05-InternVL-2.5/}{\textcolor{color_1}{InternVL2.5-38B}} & \textbf{78.85} & \underline{17.12} & \underline{17.95} & \textbf{3.31} & \underline{27.08} & \underline{8.33} & \textbf{17.54} & \textbf{25.80} & \underline{2.08} \\
\href{https://huggingface.co/openbmb/MiniCPM-V-2_6}{\textcolor{color_1}{MiniCPM-V-2\_6}} & 23.26 & 0.13 & 0.03 & 0. & 1.68 & 0. & 0. & 0. & 0. \\
\href{https://huggingface.co/Qwen/Qwen2.5-VL-3B-Instruct}{\textcolor{color_1}{Qwen2.5-VL-3B}} & 11.47 & 0.84 & 0.89 & 0. & 2.11 & 0.43 & 0.57 & 0. & 0.  \\
\href{https://huggingface.co/Qwen/Qwen2.5-VL-7B-Instruct}{\textcolor{color_1}{Qwen2-VL-7B}} & 34.65 & 0.69 & 0.55 & 0. & 1.55 & 0.43 & 2.71 & 0. & 0. \\
\href{https://huggingface.co/Qwen/Qwen2.5-VL-72B-Instruct}{\textcolor{color_1}{Qwen2-VL-72B}} & \underline{77.93} & \textbf{33.87} & \textbf{40.11} & \underline{1.49} & \textbf{39.91} & \textbf{30.70} & \underline{10.27} & \underline{4.83} & \underline{2.08} \\

\rowcolor{lightgray!20}
\midrule
\multicolumn{10}{c}{\textbf{\textit{Proprietary MLLMs}}} \\
\hline
\href{https://deepmind.google/technologies/gemini/flash/}{\textcolor{color_2}{Gemini-2.0-flash-exp}}  & 72.51 & \underline{35.54} & \underline{40.70} & \underline{26.60} & 30.18 & \underline{5.70} & \underline{12.41} & 8.06 & \textbf{8.33} \\
\href{https://ai.google.dev/gemini-api/docs/thinking-mode}{\textcolor{color_2}{Gemini-2.0-thinking-exp}} & \underline{72.61} & 35.27 & 40.06 & \textbf{27.67} & \underline{32.44} & 4.82 & 11.98 & \underline{9.13} & \textbf{8.33} \\
\href{https://openai.com/index/hello-gpt-4o/}{\textcolor{color_2}{GPT-4o}} & \textbf{85.10} & \textbf{44.89} & \textbf{52.94} & 9.61 & \textbf{40.90} & \textbf{27.19} & \textbf{16.11} & \textbf{30.10} & \underline{2.08} \\

\rowcolor{lightgray!20}
\midrule
\multicolumn{10}{c}{\textbf{\textit{Fine-tuned MLLMs on ProJudge-173k}}} \\
\hline
\href{https://internvl.github.io/blog/2024-12-05-InternVL-2.5/}{\textcolor{color_3}{InternVL2.5-8B$^{\dagger}$}}& \textbf{84.50}$_{\hl{+58.92}}$ & \textbf{45.39}$_{\hl{+38.62}}$ & \textbf{53.97}$_{\hl{+45.78}}$ & \underline{30.98}$_{\hl{+30.98}}$ & 22.14$_{\hl{+13.96}}$ & \underline{8.33}$_{\hl{+6.58}}$ & \textbf{15.12}$_{\hl{+13.95}}$ & \underline{3.22}$_{\hl{+2.15}}$ & \textbf{10.41}$_{\hl{+10.41}}$ \\
\href{https://huggingface.co/Qwen/Qwen2.5-VL-3B-Instruct}{\textcolor{color_3}{Qwen2.5-VL-3B$^{\dagger}$}}& 81.29$_{\hl{+70.45}}$ & 39.09$_{\hl{+38.25}}$ & 46.36$_{\hl{+45.47}}$ & 24.67$_{\hl{+24.67}}$ & \underline{24.54}$_{\hl{+22.43}}$ & 5.70$_{\hl{+5.27}}$ & 12.69$_{\hl{+12.12}}$ & 1.61$_{\hl{+1.61}}$ & 4.16$_{\hl{+4.16}}$ \\
\href{https://huggingface.co/Qwen/Qwen2.5-VL-7B-Instruct}{\textcolor{color_3}{Qwen2-VL-7B$^{\dagger}$}}& \underline{83.72}$_{\hl{+49.07}}$ & \underline{44.57}$_{\hl{+43.88}}$ & \underline{52.29}$_{\hl{+51.47}}$ & \textbf{31.62}$_{\hl{+31.62}}$ & \textbf{27.36}$_{\hl{+25.81}}$ & \textbf{11.40}$_{\hl{+10.97}}$ & \underline{14.69}$_{\hl{+11.98}}$ & \textbf{4.30}$_{\hl{+4.30}}$ & \underline{6.25}$_{\hl{+6.25}}$ \\

    \bottomrule
    \end{tabular}
  }
  \vspace{-1mm}
    \caption{Performances comparison of MLLMs on ProJudgeBench. The best performance is in \textbf{bold}, while the second-best is \underline{underlined}. ${\dagger}$ denotes the models fine-tuned on ProJudge-173k with Dynamic Dual-Phase strategy. In fine-tuned MLLMs, ${\hl{+}}$
    (or $\hlr{-}$) 
    indicates the performance gain
    (or lose) 
    compared to the baseline open-source models. 
  }
  \label{tab:main}
  \vspace{-4mm}
\end{table*}
\section{Experiments}
\label{sec:experiments_setup}
To provide a comprehensive evaluation of various models on ProJudgeBench, we select a diverse set of both proprietary and open-source MLLMs, including GPT-4o~\cite{ref:gpt4o}, Gemini-2.0-flash-exp~\cite{ref:gemini_2_flash}, Gemini-2.0-thinking-exp~\cite{ref:gemini_2_flash_thinking}, InternVL2.5~\cite{ref:internvl2.5} (8B, 26B, 38B), MiniCPM-V-2\_6~\cite{ref:minicpm} (8B), and Qwen2.5-VL-Instruct~\cite{ref:qwen2.5-vl} (3B, 7B, 72B). Additionally, we evaluate fine-tuned versions of InternVL2.5-8B, Qwen2.5-VL-3B, and Qwen2.5-VL-7B.

To ensure consistency, all models use the same input format and evaluation prompts, with hyperparameters following configurations in VLMEvalKit~\cite{ref:vlmevalkit}.
The evaluation prompts are provided in Appendix~\ref{appendix_prompts_process_evaluation}.
We use accuracy as the primary evaluation metric, focusing on two aspects: (1) Step Correctness Accuracy, the accuracy of determining whether each step in a student's solution is correct; and (2) Error Types Classification Accuracy, the accuracy of identifying specific error types, 
including Reasoning Errors (RE), Visual Interpretation Errors (VI), Numerical Calculation Errors (NC), Symbolic Calculation Errors (SC), Knowledge Errors (KE), Question Understanding Errors (QU), and No Solution Provided (NS).

\begin{figure*}[!t]
\centering
  \resizebox{0.9\linewidth}{!} {
    \includegraphics{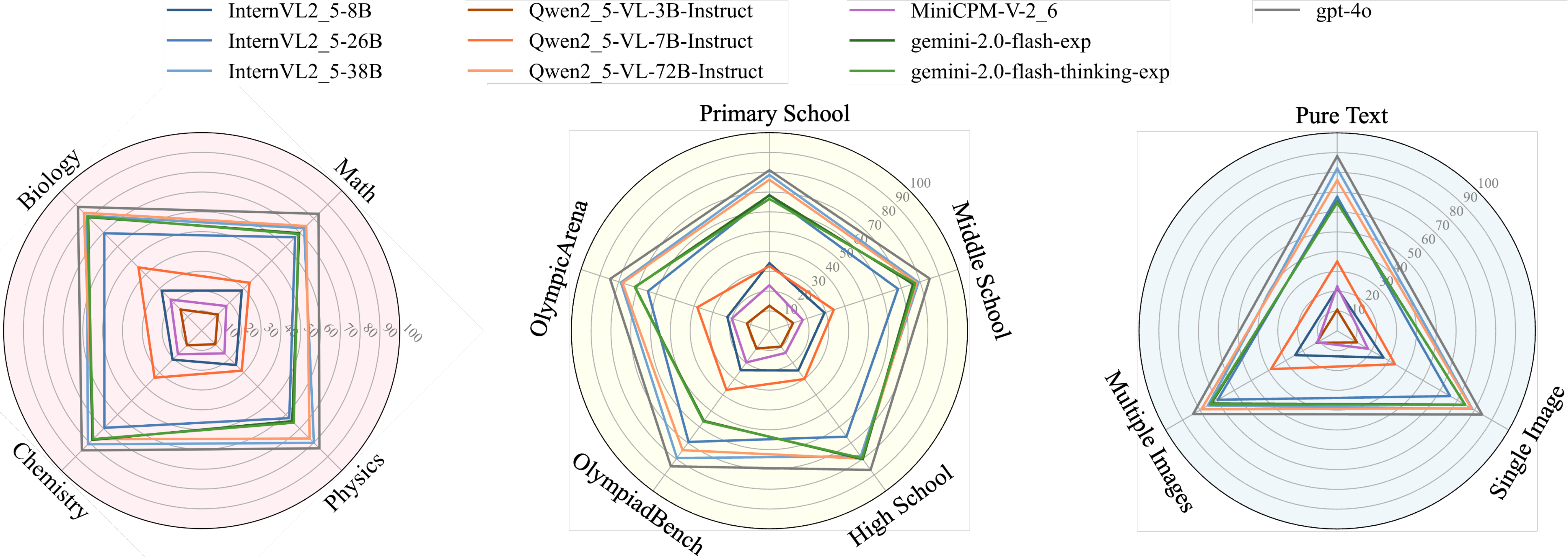}
  }
  \caption{Performance of MLLM-based process judges across different disciplines, difficulty levels and modalities.}
    \label{fig:radar}
  \vspace{-5mm}
\end{figure*}
\subsection{Main Results and Findings}
In this subsection, we compare the performance of 11 open-source and proprietary models and 3 fine-tuned models. The experimental results are presented in Table~\ref{tab:main}. 

\noindent \textbf{Finding 1: Model Scale Impacts Process Evaluation Capabilities.}  
Larger models exhibit significantly higher accuracies in both step correctness and error type classification.
For example, InternVL series exhibits a clear scaling effect, with the step correctness accuracy increasing from 25.58\% for the 8B model to 66.72\% and 78.85\% for the 26B and 38B versions, respectively. 
In contrast, smaller models like MiniCPM-V-2\_6 exhibit weaker reasoning capabilities, with near-zero accuracy in several error categories.

\noindent \textbf{Finding 2: Proprietary Models Maintain a Performance Lead.}
GPT-4o achieves the highest step correctness accuracy and excels across multiple error categories. Gemini-2.0 series, while slightly behind GPT-4o, still outperforms most open-source models, with step correctness accuracy around 72.5\% and shows particular strength in identifying visual interpretation errors. These results underscore the advanced multi-modal reasoning capabilities of proprietary models. 

\noindent \textbf{Finding 3: Fine-tuning Significantly Enhances Model Performance.}
Open-source models like InternVL2.5-8B and Qwen2.5-VL-3B show significant gains after fine-tuning, with step correctness accuracy increasing by 58.92\% and 70.45\% respectively, bringing their performance on par with proprietary models.
With sufficient fine-tuning, smaller models can approach the performance of much larger counterparts. For instance, the fine-tuned Qwen2.5-VL-3B achieves 81.29\% accuracy, surpassing Qwen2.5-VL-72B.
These results highlight the transformative impact of domain-specific fine-tuning.

\noindent \textbf{Finding 4: Models Exhibit Unique Strengths Across Different Error Types.}
While GPT-4o leads overall, open source models like Qwen2.5-VL-72B and InternVL2.5-38B also excel in certain tasks. For example, Qwen2.5-VL-72B performs remarkbly well in identifying reasoning and calculation errors, while InternVL2.5-38B leads in detecting knowledge and question understanding errors, highlighting the nuanced strengths of different model architectures.


\subsection{Further Analysis}

To gain deeper insights into the capabilities of MLLMs as process judges, we explore several key research questions~(RQ): (1) How does model performance vary across different domains? (2) How do models perform as judges when assessing solutions from different student models? (3) What is the relationship between a model’s ability to detect errors and its tendency to generate similar errors?

\subsubsection{RQ1: Performance Across Different Domains}
We analyze model performance across different disciplines (math, physics, chemistry, biology), difficulty levels (primary, middle, high school, competition), and modalities (pure text, single image, multiple images). The results, illustrated in Figure~\ref{fig:radar}, reveal several key insights into how MLLM-based process judges perform in diverse scenarios.

\noindent \textbf{Most MLLMs exhibit higher accuracy in biology and chemistry, compared to mathematics and physics.} 
For instance, Qwen2.5-VL-72B achieves 84.13\% accuracy in biology but only 77.08\% in physics, while InternVL2.5-38B achieves 81.22\% in chemistry but 73.14\% in math.
This discrepancy can be attributed to the nature of the tasks: biology and chemistry often rely on factual knowledge and rule-based reasoning, making them easier for models to evaluate.
In contrast, math and physics demand complex logical reasoning, multi-step derivations and the ability to interprete real-world scenarios, which require precise logical validation during evaluation, leading to misjudgements when models fail to grasp subtle errors in reasoning.

\noindent \textbf{As task difficulty increases, model accuracy in step evaluation declines.} 
For instance, Gemini-2.0 achieves 80.2\% accuracy in high school tasks but drops to 56.2\% in OlympiadBench tasks.
This trend reflects the challenges for current models to deal with complex, multi-step reasoning tasks, which require both advanced domain knowledge and the ability to synthesize information in novel ways.
Interestingly, primary school tasks do not always show the highest accuracy, despite their relative simplicity. 
This can be attributed to their variablity and context-dependent nature, which may involve unconventional reasoning steps or less structured problem-solving processes, posing challengies for models trained on formal and structured datasets.

\noindent \textbf{Open-source models face challenges with tasks involving images.} 
Pure text tasks, which rely solely on textual information, align well with the core strengths of language models, resulting in more accurate step evaluation. In contrast, tasks involving images require models to interpret visual elements and integrate them with textural reasoning, complicating the evaluation process and reducing accuracy. 
This underscores the limitations of current open-source models in multi-modal reasoning and evaluation.


\begin{figure}[!t]
\centering
  \resizebox{0.95\linewidth}{!} {
    \includegraphics{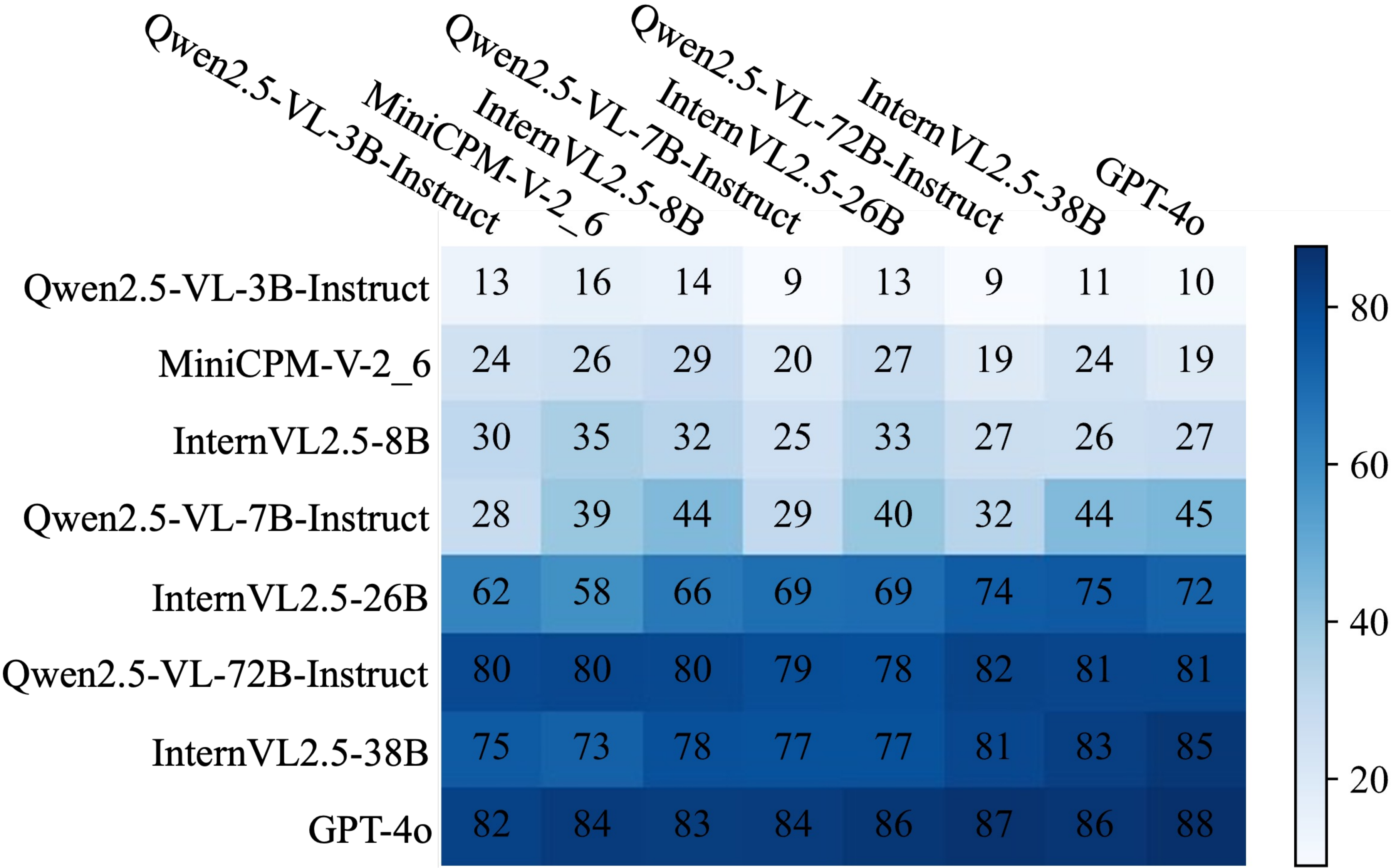}
  }
  \caption{Model-as-Judge Performance across different models.
  Each position (x, y) in the heatmap represents the accuracy of model x as a judge in assessing solutions generated by model y.
  }
    \label{fig:heatmap}
  \vspace{-5.5mm}
\end{figure}
\begin{figure}[!t]
\centering
  \resizebox{0.9\linewidth}{!} {
    \includegraphics{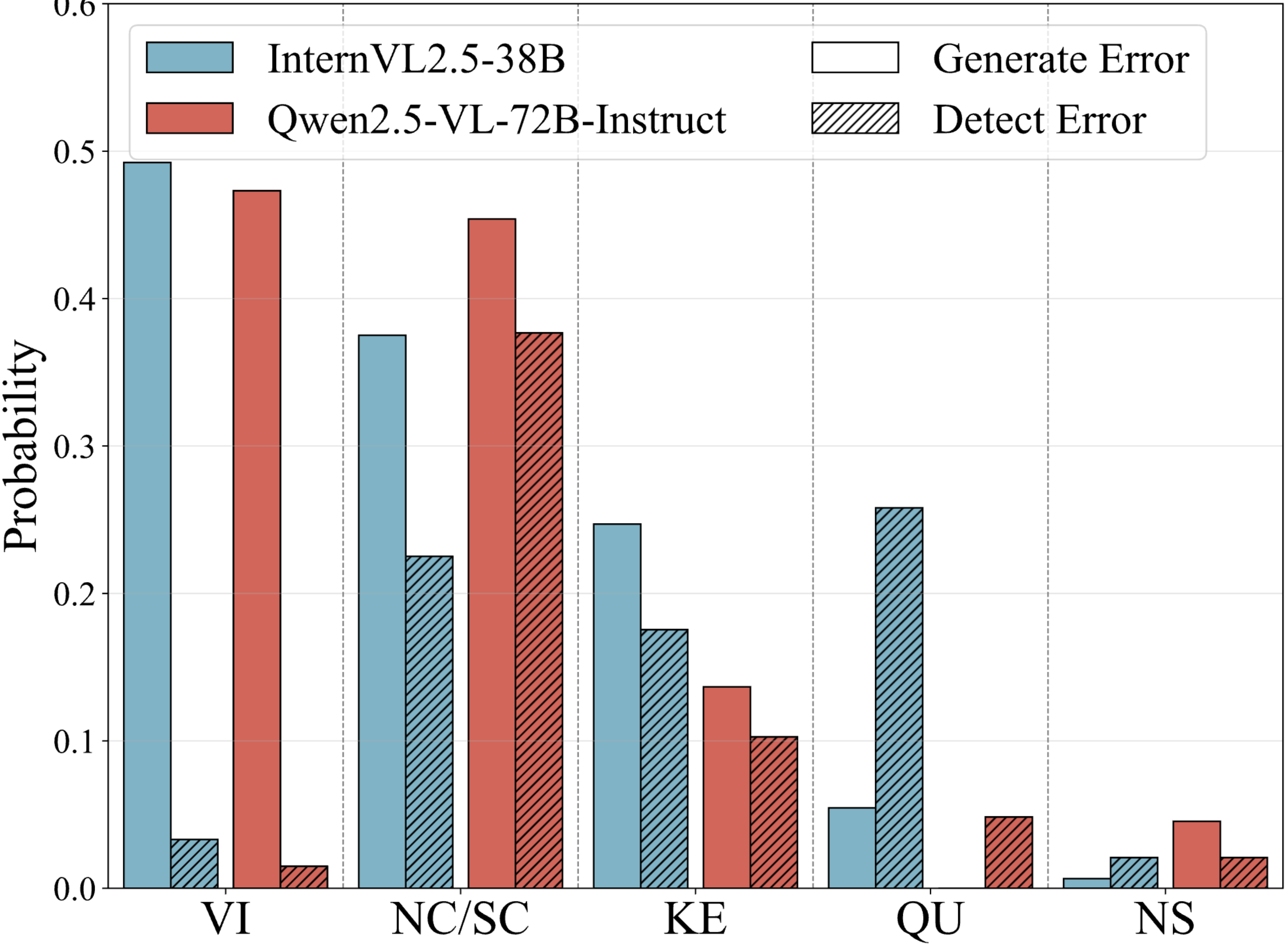}
  }
  \vspace{-2mm}
  \caption{
  Relationship between error generation and detection. 
  VI, NC/SC, KE, QU and NS represent for visual interpretation errors, calculation errors, knowledge errors, question understanding errors and no solution provided respectively.
  }
    \label{fig:bar_making_and_identifying_error}
  \vspace{-6mm}
\end{figure}
\subsubsection{RQ2: Cross-Model Evalution}
Figure~\ref{fig:heatmap} illustrates the performance of different models acting as process judges when evaluating solutions generated by other models (student models). 
We find that InternVL models demonstrate higher accuracy when evaluating themselves or models from the same series.
This can be attributed to the shared architectures, training data, and reasoning styles within the same series, enabling them to better understand and assess the reasoning processes and error patterns of their counterparts.
Besides, larger models tend to achieve higher accuracy when evaluating smaller models, whereas smaller models struggle to assess larger ones effectively.
For example, GPT-4o reaches 84.31\% accuracy when evaluating MiniCPM-V-2\_6, while MiniCPM-V-2\_6 achieves only 15.69\% when evaluating GPT-4o. 
This indicates the superior reasoning abilities of larger models, which enable them to better 
analyse and validate the outputs of smaller models.
In contrast, smaller models, constrained by their limited capacity, often struggle to accurately evaluate the more complex reasoning steps of larger models.

\subsubsection{RQ3: Error Detection vs. Error Generation}
Given that models vary in their ability to detect different error types, we aim to explore the relationship between error types and detection performance.
We investigate two MLLMs across five error types, examining: their frequency of generating these errors when acting as student models; and their accuracy in detecting these errors as process judges.
The results are shown in Figure~\ref{fig:bar_making_and_identifying_error}.

Both models exhibit high generation rates and low detection accuracy in visual interpretation errors, revealing their limitations in 
interpreting and reasoning about visual information.
For calculation errors, both models have a relatively high generation frequency and also notable detection accuracy, indicating that while models are prone to calculation errors, they possess a reasonable ability to identify such errors. This suggests the potential for reducing such errors by introducing self-checking mechanisms.
Besides, both models show low knowledge error generation frequency and detection accuracy, suggesting that models may lack domain knowledge in certain areas, leading to mistakes when encountering related concepts and an inability to identify such errors. Targeted training on relevant domain knowledge could help improve performance in both aspects.

\begin{table}[t]
\centering
\resizebox{0.4\textwidth}{!}{
\begin{tabular}{lcccc}
\toprule
\textbf{Model} & \textbf{Overall} & \textbf{OlympicArena} \\
\midrule
InternVL2.5-8B (Base) & 25.58 & 22.33 \\
+ FT & 83.37 & 81.71 \\
+ FT + DDP (Full) & \textbf{84.50} & \textbf{85.07} \\
\midrule
QwenVL2.5-3B (Base) & 11.47 & 11.92 \\ 
+ FT & 80.57 & 80.38 \\
+ FT + DDP (Full) & \textbf{81.29} & \textbf{81.01} \\
\midrule 
QwenVL2.5-7B (Base) & 34.65 & 38.25 \\
+ FT & 81.91 & 77.88 \\
+ FT + DDP (Full) & \textbf{83.72} & \textbf{82.67} \\
\bottomrule
\end{tabular}
}
\vspace{-1.5mm}
\caption{Ablation study on Dynamic Dual-Phase fine-tuning strategy. 
We compare the standard fine-tuning (+ FT) with DDP-enhanced training (+ DDP).
Overall and OlympicArena represent the results on the full set of ProJudgeBench and its OlympicArena subset respectively.
}
\label{tab:ablation}
\vspace{-4.78mm}
\end{table}

\subsection{Ablation Study}
In this section, we conduct an ablation study to evaluate the effectiveness of our Dynamic Dual-Phase (DDP) fine-tuning strategy. As shown in Table \ref{tab:ablation}, we compare the standard fine-tuning (+FT) with DDP-enhanced training (+DDP). 
Our evaluation focuses on two aspects: overall step correctness accuracy and results on the unseen OlympicArena test set, which allows us to assess both in-domain performance and generalization capabilities.
All models were fine-tuned for one epoch using LoRA~\cite{ref:lora}, with further details provided in Appendix~\ref{appendix_fine_tuning_details}.

The results demonstrate that DDP strategy consistently improves performance across all models, with particularly notable gains in generalization to unseen problem types. 
On the challenging OlymicArena test set, which contains entirely novel competition-level problems, DDP-enhanced models achieve significant performance boosts—InternVL2.5-8B improves by 3.36\%, while QwenVL2.5-7B gains 4.79\%. This suggests that the explicit problem-solving phase in DDP not only deepens the model's understanding of scientific reasoning, but also enhances its robustness and generalizability to evaluate unfamiliar solutions.

\section{Conclusions}

In this paper, we introduce ProJudgeBench, a multi-modal, multi-discipline benchmark, assessing the fine-grained error detection, classification and diagnosis capabilities of MLLM-based process judges.
We also present ProJudge-173k, a large-scale instruction-tuning dataset, and a Dynamic Dual-Phase fine-tuning strategy. Both significantly enhance the process evaluation capabilities of open-source models.
Through extensive experiments, we identify key challenges in current models and provide actionable insights for future advancements. 
Our work lays a foundation for process evaluation in multi-modal reasoning, and we hope it will inspire future research in this critical area.

{
    \small
    \bibliographystyle{unsrt} 
    \bibliography{main}

\begin{thebibliography}{10}

\bibitem{ref:gpt4o}
OpenAI.
\newblock Gpt-4o system card.
\newblock \url{https://cdn.openai.com/gpt-4o-system-card.pdf}, 2024.
\newblock Accessed: 2024-09-26.

\bibitem{ref:o1}
OpenAI.
\newblock Learning to reason with llms.
\newblock \url{https://openai.com/index/learning-to-reason-with-llms/}, September 2024.

\bibitem{ref:gemini_2_flash}
{DeepMind}.
\newblock Gemini 2.0 flash experimental.
\newblock \url{https://deepmind.google/technologies/gemini/flash/}, 2024.
\newblock Accessed: 2024-12-25.

\bibitem{ref:qvq}
Qwen Team.
\newblock Qvq: To see the world with wisdom, December 2024.

\bibitem{ref:mmmu}
Xiang Yue, Yuansheng Ni, Kai Zhang, Tianyu Zheng, Ruoqi Liu, Ge~Zhang, Samuel Stevens, Dongfu Jiang, Weiming Ren, Yuxuan Sun, et~al.
\newblock Mmmu: A massive multi-discipline multimodal understanding and reasoning benchmark for expert agi.
\newblock In {\em Proceedings of the IEEE/CVF Conference on Computer Vision and Pattern Recognition}, pages 9556--9567, 2024.

\bibitem{ref:gsm8k}
Karl Cobbe, Vineet Kosaraju, Mohammad Bavarian, Mark Chen, Heewoo Jun, Lukasz Kaiser, Matthias Plappert, Jerry Tworek, Jacob Hilton, Reiichiro Nakano, Christopher Hesse, and John Schulman.
\newblock Training verifiers to solve math word problems.
\newblock {\em arXiv preprint arXiv:2110.14168}, 2021.

\bibitem{ref:math}
Dan Hendrycks, Collin Burns, Saurav Kadavath, Akul Arora, Steven Basart, Eric Tang, Dawn Song, and Jacob Steinhardt.
\newblock Measuring mathematical problem solving with the math dataset.
\newblock In {\em Proceedings of the Neural Information Processing Systems Track on Datasets and Benchmarks}, volume~1, 2021.

\bibitem{ref:olympicarena}
Zhen Huang, Zengzhi Wang, Shijie Xia, Xuefeng Li, Haoyang Zou, Ruijie Xu, Run-Ze Fan, Lyumanshan Ye, Ethan Chern, Yixin Ye, et~al.
\newblock Olympicarena: Benchmarking multi-discipline cognitive reasoning for superintelligent ai.
\newblock {\em Advances in Neural Information Processing Systems}, 37:19209--19253, 2025.

\bibitem{ref:reasoneval}
Shijie Xia, Xuefeng Li, Yixin Liu, Tongshuang Wu, and Pengfei Liu.
\newblock Evaluating mathematical reasoning beyond accuracy.
\newblock {\em arXiv preprint arXiv:2404.05692}, 2024.

\bibitem{ref:uesato2022solving}
Jonathan Uesato, Nate Kushman, Ramana Kumar, Francis Song, Noah Siegel, Lisa Wang, Antonia Creswell, Geoffrey Irving, and Irina Higgins.
\newblock Solving math word problems with process-and outcome-based feedback.
\newblock {\em arXiv preprint arXiv:2211.14275}, 2022.

\bibitem{ref:prm800k}
Hunter Lightman, Vineet Kosaraju, Yura Burda, Harri Edwards, Bowen Baker, Teddy Lee, Jan Leike, John Schulman, Ilya Sutskever, and Karl Cobbe.
\newblock Let's verify step by step.
\newblock {\em arXiv preprint arXiv:2305.20050}, 2023.

\bibitem{ref:mr_gsm8k}
Zhongshen Zeng, Pengguang Chen, Shu Liu, Haiyun Jiang, and Jiaya Jia.
\newblock Mr-gsm8k: A meta-reasoning benchmark for large language model evaluation.
\newblock {\em arXiv preprint arXiv:2312.17080}, 2023.

\bibitem{ref:criticbench}
Zicheng Lin, Zhibin Gou, Tian Liang, Ruilin Luo, Haowei Liu, and Yujiu Yang.
\newblock {C}ritic{B}ench: Benchmarking {LLM}s for critique-correct reasoning.
\newblock In {\em Findings of the Association for Computational Linguistics: ACL 2024}, pages 1552--1587, 2024.

\bibitem{ref:mathcheck_gsm}
Zihao Zhou, Shudong Liu, Maizhen Ning, Wei Liu, Jindong Wang, Derek~F Wong, Xiaowei Huang, Qiufeng Wang, and Kaizhu Huang.
\newblock Is your model really a good math reasoner? evaluating mathematical reasoning with checklist.
\newblock {\em arXiv preprint arXiv:2407.08733}, 2024.

\bibitem{ref:processbench}
Chujie Zheng, Zhenru Zhang, Beichen Zhang, Runji Lin, Keming Lu, Bowen Yu, Dayiheng Liu, Jingren Zhou, and Junyang Lin.
\newblock Processbench: Identifying process errors in mathematical reasoning.
\newblock {\em arXiv preprint arXiv:2412.06559}, 2024.

\bibitem{ref:prmbench}
Mingyang Song, Zhaochen Su, Xiaoye Qu, Jiawei Zhou, and Yu~Cheng.
\newblock Prmbench: A fine-grained and challenging benchmark for process-level reward models.
\newblock {\em arXiv preprint arXiv:2501.03124}, 2025.

\bibitem{ref:mm-math}
Kai Sun, Yushi Bai, Ji~Qi, Lei Hou, and Juanzi Li.
\newblock {MM}-{MATH}: Advancing multimodal math evaluation with process evaluation and fine-grained classification.
\newblock In {\em Findings of the Association for Computational Linguistics: EMNLP 2024}, pages 1358--1375, 2024.

\bibitem{ref:mathverse}
Renrui Zhang, Dongzhi Jiang, Yichi Zhang, Haokun Lin, Ziyu Guo, Pengshuo Qiu, Aojun Zhou, Pan Lu, Kai-Wei Chang, Peng Gao, et~al.
\newblock Mathverse: Does your multi-modal llm truly see the diagrams in visual math problems?
\newblock {\em arXiv preprint arXiv:2403.14624}, 2024.

\bibitem{ref:mme-cot}
Dongzhi Jiang, Renrui Zhang, Ziyu Guo, Yanwei Li, Yu~Qi, Xinyan Chen, Liuhui Wang, Jianhan Jin, Claire Guo, Shen Yan, et~al.
\newblock Mme-cot: Benchmarking chain-of-thought in large multimodal models for reasoning quality, robustness, and efficiency.
\newblock {\em arXiv preprint arXiv:2502.09621}, 2025.

\bibitem{ref:qwen2.5-vl}
Qwen Team.
\newblock Qwen2.5-vl, January 2025.

\bibitem{ref:blip}
Junnan Li, Dongxu Li, Caiming Xiong, and Steven Hoi.
\newblock Blip: Bootstrapping language-image pre-training for unified vision-language understanding and generation.
\newblock In {\em International conference on machine learning}, pages 12888--12900. PMLR, 2022.

\bibitem{ref:deepseekvl}
Haoyu Lu, Wen Liu, Bo~Zhang, Bingxuan Wang, Kai Dong, Bo~Liu, Jingxiang Sun, Tongzheng Ren, Zhuoshu Li, Hao Yang, et~al.
\newblock Deepseek-vl: towards real-world vision-language understanding.
\newblock {\em arXiv preprint arXiv:2403.05525}, 2024.

\bibitem{ref:mathvista}
Pan Lu, Hritik Bansal, Tony Xia, Jiacheng Liu, Chunyuan Li, Hannaneh Hajishirzi, Hao Cheng, Kai-Wei Chang, Michel Galley, and Jianfeng Gao.
\newblock Mathvista: Evaluating mathematical reasoning of foundation models in visual contexts.
\newblock In {\em The Twelfth International Conference on Learning Representations}.

\bibitem{ref:liu2024mm}
Xin Liu, Yichen Zhu, Jindong Gu, Yunshi Lan, Chao Yang, and Yu~Qiao.
\newblock Mm-safetybench: A benchmark for safety evaluation of multimodal large language models.
\newblock In {\em European Conference on Computer Vision}, pages 386--403. Springer, 2024.

\bibitem{ref:eee-bench}
Ming Li, Jike Zhong, Tianle Chen, Yuxiang Lai, and Konstantinos Psounis.
\newblock Eee-bench: A comprehensive multimodal electrical and electronics engineering benchmark.
\newblock {\em arXiv preprint arXiv:2411.01492}, 2024.

\bibitem{ref:lvlm}
Peng Xu, Wenqi Shao, Kaipeng Zhang, Peng Gao, Shuo Liu, Meng Lei, Fanqing Meng, Siyuan Huang, Yu~Qiao, and Ping Luo.
\newblock Lvlm-ehub: A comprehensive evaluation benchmark for large vision-language models.
\newblock {\em IEEE Transactions on Pattern Analysis and Machine Intelligence}, 2024.

\bibitem{ref:mmbench}
Yuan Liu, Haodong Duan, Yuanhan Zhang, Bo~Li, Songyang Zhang, Wangbo Zhao, Yike Yuan, Jiaqi Wang, Conghui He, Ziwei Liu, et~al.
\newblock Mmbench: Is your multi-modal model an all-around player?
\newblock In {\em European conference on computer vision}, pages 216--233. Springer, 2024.

\bibitem{ref:mmt-bench}
Kaining Ying, Fanqing Meng, Jin Wang, Zhiqian Li, Han Lin, Yue Yang, Hao Zhang, Wenbo Zhang, Yuqi Lin, Shuo Liu, et~al.
\newblock Mmt-bench: A comprehensive multimodal benchmark for evaluating large vision-language models towards multitask agi.
\newblock In {\em Forty-first International Conference on Machine Learning}.

\bibitem{masry2022chartqa}
Ahmed Masry, Do~Xuan Long, Jia~Qing Tan, Shafiq Joty, and Enamul Hoque.
\newblock Chartqa: A benchmark for question answering about charts with visual and logical reasoning.
\newblock {\em arXiv preprint arXiv:2203.10244}, 2022.

\bibitem{ref:seed-bench}
Bohao Li, Yuying Ge, Yixiao Ge, Guangzhi Wang, Rui Wang, Ruimao Zhang, and Ying Shan.
\newblock Seed-bench: Benchmarking multimodal large language models.
\newblock In {\em Proceedings of the IEEE/CVF Conference on Computer Vision and Pattern Recognition}, pages 13299--13308, 2024.

\bibitem{ref:ok-bench}
Kenneth Marino, Mohammad Rastegari, Ali Farhadi, and Roozbeh Mottaghi.
\newblock Ok-vqa: A visual question answering benchmark requiring external knowledge.
\newblock In {\em Proceedings of the IEEE/cvf conference on computer vision and pattern recognition}, pages 3195--3204, 2019.

\bibitem{ref:docvqa}
Minesh Mathew, Dimosthenis Karatzas, and CV~Jawahar.
\newblock Docvqa: A dataset for vqa on document images.
\newblock In {\em Proceedings of the IEEE/CVF winter conference on applications of computer vision}, pages 2200--2209, 2021.

\bibitem{ref:wemath}
Runqi Qiao, Qiuna Tan, Guanting Dong, Minhui Wu, Chong Sun, Xiaoshuai Song, Zhuoma GongQue, Shanglin Lei, Zhe Wei, Miaoxuan Zhang, et~al.
\newblock We-math: Does your large multimodal model achieve human-like mathematical reasoning?
\newblock {\em arXiv preprint arXiv:2407.01284}, 2024.

\bibitem{qiao2024we}
Runqi Qiao, Qiuna Tan, Guanting Dong, Minhui Wu, Chong Sun, Xiaoshuai Song, Zhuoma GongQue, Shanglin Lei, Zhe Wei, Miaoxuan Zhang, et~al.
\newblock We-math: Does your large multimodal model achieve human-like mathematical reasoning?
\newblock {\em arXiv preprint arXiv:2407.01284}, 2024.

\bibitem{ref:math-vision}
Ke~Wang, Junting Pan, Weikang Shi, Zimu Lu, Houxing Ren, Aojun Zhou, Mingjie Zhan, and Hongsheng Li.
\newblock Measuring multimodal mathematical reasoning with math-vision dataset.
\newblock {\em Advances in Neural Information Processing Systems}, 37:95095--95169, 2025.

\bibitem{wang2025charxiv}
Zirui Wang, Mengzhou Xia, Luxi He, Howard Chen, Yitao Liu, Richard Zhu, Kaiqu Liang, Xindi Wu, Haotian Liu, Sadhika Malladi, et~al.
\newblock Charxiv: Charting gaps in realistic chart understanding in multimodal llms.
\newblock {\em Advances in Neural Information Processing Systems}, 37:113569--113697, 2025.

\bibitem{ref:gpt4v}
OpenAI.
\newblock Gpt-4v(ision) system card.
\newblock \url{https://cdn.openai.com/papers/GPTV_System_Card.pdf}, 2023.
\newblock Accessed: 2024-09-26.

\bibitem{ref:math_shepherd}
Peiyi Wang, Lei Li, Zhihong Shao, RX~Xu, Damai Dai, Yifei Li, Deli Chen, Y~Wu, and Zhifang Sui.
\newblock Math-shepherd: Verify and reinforce llms step-by-step without human annotations.
\newblock {\em CoRR, abs/2312.08935}, 2023.

\bibitem{ref:olympiadbench}
Chaoqun He, Renjie Luo, Yuzhuo Bai, Shengding Hu, Zhen Thai, Junhao Shen, Jinyi Hu, Xu~Han, Yujie Huang, Yuxiang Zhang, Jie Liu, Lei Qi, Zhiyuan Liu, and Maosong Sun.
\newblock {O}lympiad{B}ench: A challenging benchmark for promoting {AGI} with olympiad-level bilingual multimodal scientific problems.
\newblock In {\em Proceedings of the 62nd Annual Meeting of the Association for Computational Linguistics}, pages 3828--3850, 2024.

\bibitem{ref:internvl2.5}
Zhe Chen, Weiyun Wang, Yue Cao, Yangzhou Liu, Zhangwei Gao, Erfei Cui, Jinguo Zhu, Shenglong Ye, Hao Tian, Zhaoyang Liu, et~al.
\newblock Expanding performance boundaries of open-source multimodal models with model, data, and test-time scaling.
\newblock {\em arXiv preprint arXiv:2412.05271}, 2024.

\bibitem{ref:llava}
Bo~Li, Yuanhan Zhang, Dong Guo, Renrui Zhang, Feng Li, Hao Zhang, Kaichen Zhang, Yanwei Li, Ziwei Liu, and Chunyuan Li.
\newblock Llava-onevision: Easy visual task transfer.
\newblock {\em arXiv preprint arXiv:2408.03326}, 2024.

\bibitem{ref:minicpm}
Yuan Yao, Tianyu Yu, Ao~Zhang, Chongyi Wang, Junbo Cui, Hongji Zhu, Tianchi Cai, Haoyu Li, Weilin Zhao, Zhihui He, et~al.
\newblock Minicpm-v: A gpt-4v level mllm on your phone.
\newblock {\em arXiv preprint arXiv:2408.01800}, 2024.

\bibitem{ref:qwen2.5}
Qwen Team.
\newblock Qwen2.5: A party of foundation models, September 2024.

\bibitem{ref:camel}
Guohao Li, Hasan Abed Al~Kader Hammoud, Hani Itani, Dmitrii Khizbullin, and Bernard Ghanem.
\newblock Camel: Communicative agents for "mind" exploration of large language model society.
\newblock In {\em Thirty-seventh Conference on Neural Information Processing Systems}, 2023.

\bibitem{ref:gemini_2_flash_thinking}
{DeepMind}.
\newblock Gemini 2.0 flash thinking.
\newblock \url{https://deepmind.google/technologies/gemini/flash-thinking/}, 2025.
\newblock Accessed: 2025-01-21.

\bibitem{ref:vlmevalkit}
Haodong Duan, Junming Yang, Yuxuan Qiao, Xinyu Fang, Lin Chen, Yuan Liu, Xiaoyi Dong, Yuhang Zang, Pan Zhang, Jiaqi Wang, et~al.
\newblock Vlmevalkit: An open-source toolkit for evaluating large multi-modality models.
\newblock In {\em Proceedings of the 32nd ACM International Conference on Multimedia}, pages 11198--11201, 2024.

\bibitem{ref:lora}
Edward~J Hu, Yelong Shen, Phillip Wallis, Zeyuan Allen-Zhu, Yuanzhi Li, Shean Wang, Lu~Wang, and Weizhu Chen.
\newblock Lo{RA}: Low-rank adaptation of large language models.
\newblock In {\em International Conference on Learning Representations}, 2022.

\end{thebibliography}
}

\newpage
\appendix
\onecolumn
\setcounter{page}{1}





\section{Related Work}
\label{appendix_related_work}
\subsection{Multi-Modal Benchmarks}
With the fast development of recent MLLMs~\cite{ref:gemini_2_flash,ref:gpt4o,ref:qwen2.5-vl,ref:blip,ref:deepseekvl}, it becomes essential to evaluate their performance across a variety of tasks to understand their strengths and weaknesses, which is crucial for guiding future developments and enhancements~\cite{ref:mmmu,ref:mathvista,ref:eee-bench,ref:mmbench,ref:mmt-bench,masry2022chartqa,ref:seed-bench,ref:ok-bench,ref:docvqa,ref:wemath,qiao2024we,ref:math-vision,wang2025charxiv}. For example, MMMU~\cite{ref:mmmu} has been proposed to evaluate the scientific problem-solving abilities of MLLMs across a range of college-level subjects. MathVista~\cite{ref:mathvista} is widely used to assess the mathematical capabilities of MLLMs. While considerable progress has been achieved, these benchmarks often focus solely on evaluating final answers, overlooking crucial information of intermediate reasoning steps and potentially leading to unreliable results.
\subsection{MLLM-based Process Judges}
As MLLMs regularly make mistakes when solving scientific problems~\cite{ref:mmmu,ref:mathverse}, evaluating the validity of their reasoning processes is critical for ensuring reliability and uncovering fine-grained model weaknesses. Since human evaluation is costly and time-consuming, prompting MLLMs as automated process judges has become a common practice.
MLLM-based process judges have been widely used to automatically evaluate reasoning steps of multi-modal language models.
For example, 
MM-Math~\cite{ref:mm-math} incorporates MLLM-as-a-judge to automatically analyze solution steps, identifying and categorizing errors into specific error types. 
OlympicArena~\cite{ref:olympicarena} employs GPT-4V to conduct process-level evaluations, scoring the correctness of each reasoning step to ensure a rigorous assessment.
MathVerse~\cite{ref:mathverse} introduces a Chain-of-Thought (CoT) evaluation strategy, using GPT-4V to extract and assess key reasoning steps, providing fine-grained error analysis and nuanced scoring that goes beyond binary correctness.
MME-CoT~\cite{ref:mme-cot} extends this approach by evaluating CoT reasoning across multiple domains, while introducing metrics for reasoning quality, robustness, and efficiency.

Despite the widespread adoption, reliability of MLLM-based judges themselves is rarely scrutinized. 
Besides, current evaluation predominantly relies on proprietary models, suffering from prohibitive costs and reproducibility instability. This necessitates the development of open-source process judges. 
To address these challenges, we introduce \textbf{ProJudgeBench}, a comprehensive benchmark for assessing MLLMs' capabilities as process judges, and \textbf{ProJudge-173k}, a large-scale instruction-tuning dataset designed to enhance open-source MLLMs' process evaluation abilities.

\subsection{Process Evaluation Benchmarks}
There exist several benchmarks related to assessing process evaluation abilities of LLMs. 
MR-GSM8K~\cite{ref:mr_gsm8k} introduces a meta-reasoning paradigm, requiring LLMs to transition from solving problems to evaluating the correctness of reasoning steps.
MathCheck-GSM~\cite{ref:mathcheck_gsm} presents a checklist-based framework, where LLMs are tasked with evaluating both final answers and intermediate reasoning steps.
CriticBench~\cite{ref:criticbench} assesses the ability of LLMs to critique and correct their reasoning across multiple domains.
ProcessBench~\cite{ref:processbench} measures the ability to identify erroneous steps in mathematical reasoning, particularly in competition-level problems.
PRMBench~\cite{ref:prmbench} synthesize erroneous steps based on PRM800k~\cite{ref:prm800k}, evaluating fine-grained error detection capabilities of PRMs across multiple dimensions.

Our ProJudgeBench is distinguished from prior benchmarks in three key aspects: 
First, it covers four scientific discipline with multi-modal content and varying difficutly levels, reflecting the complexity of real-world reasoning tasks.
Second, test cases are curated from diverse model-generated solutions instead of synthetic data, capturing broad realistic reasoning behaviors and error patterns.
Third, each step is human-annotated for correctness, error type and explanation, enabling fine-grained evaluation of models' error-diagnosing capabilities.
\section{Definitions of Error Types}
\label{appendix_error_types_definition}
As described in Section~\ref{sec:task_definition}, we define seven error types based on a comprehensive analysis of common mistakes that models tend to make during long-chain reasoning processes. The definitions of each category are displayed in Table~\ref{tab:error_types_definitions}.
\begin{table*}[h!]
  \centering
  \renewcommand{\arraystretch}{1.2}
  \begin{tabularx}{0.85\textwidth}{@{}>{\RaggedRight}p{4.5cm}>{\RaggedRight}X@{}}
    \toprule
    \textbf{Error Types} & \textbf{Definitions} \\
    \midrule
    Numerical Calculation Error & Errors in basic arithmetic operations such as addition, subtraction, division, or square roots.\\
    \hline
    Symbolic Calculation Error & Errors in manipulating algebraic expressions, such as incorrect expansion, factoring, simplification, or solving equations with variables.\\
    \hline
    Visual Interpretation Error & Errors in interpreting graphical data, such as misidentifying coordinates, shapes, spatial relationships, or data within figures. \\
    \hline
    Reasoning Error & Errors in the logical thinking process that lead to incorrect conclusions, such as flawed arguments, invalid inferences, or gaps in the logical flow of the solution. \\
    \hline
    Knowledge Error & Errors caused by insufficient understanding or incorrect application of necessary knowledge (e.g., concepts, formulas, theorems, methods), or using outdated or incorrect information. \\
    \hline
    Question Understanding Error & Errors due to misunderstanding or misinterpreting the problems' conditions or requirements, such as misreading questions or misapplying given conditions. \\
    \hline
    No solution provided & The model refuses to answer, fails to follow instructions to make a solution, or encounters anomalies in generation process such as repetitive responses or incomplete outputs. \\
    \bottomrule
  \end{tabularx}
  \caption{Definitions of seven error types.}
  \label{tab:error_types_definitions}
\end{table*}
\section{Detailed Information of ProJudgeBench}
In this section, we provide detailed information on ProJudgeBench, including the list of MLLMs used for generating solutions in data construction, instruction for human annotators, our annotation website, quality control process, and breakdown statistics.

\subsection{Data Construction}
\label{appendix_data_construction_ProJudgeBench}
As described in Section~\ref{sec:ProJudgeBench_data_construction}, we utilize 10 distinct MLLMs to generate solutions in ProJudgeBench. The list of utilized MLLMs are presented in Table~\ref{tab:MLLMs_list_ProJudgeBench}.
\begin{table*}[h!]
\centering
  \resizebox{0.75\textwidth}{!}{
    \begin{tabular}{l|l|l|l}
    \toprule[0.5pt]
\multicolumn{4}{c}{\textbf{MLLMs as solution generators in ProJudgeBench}} \\
\midrule
\tabincell{l}{
InternVL2.5-8B \\
InternVL2.5-26B \\
InternVL2.5-38B
} &
\tabincell{l}{
Qwen2.5-VL-Instruct-3B \\
Qwen2.5-VL-Instruct-7B \\
Qwen2.5-VL-Instruct-72B 
} &
\tabincell{l}{
MiniCPM-V-2\_6 (8B) \\
QVQ-72B-Preview
}&
\tabincell{l}{
LLaVA-OneVision (7B) \\
GPT-4o
}
\\

    \bottomrule[0.5pt]
    \end{tabular}
  }
    \caption{List of MLLMs used in ProJudgeBench to generate solutions.
  }
  \label{tab:MLLMs_list_ProJudgeBench}
\end{table*}


\subsection{Quality Control}
\label{appendix_quality_control_ProJudgeBench}
To ensure annotation reliability, 
we implement a rigorous quality control mechanism. 
First, a random subset of solutions are selected for repeated annotation to measure inter-annotator agreement. In cases where annotations exhibit significant discrepancies, the solutions are flagged for re-annotation.
Additionally, solutions with missing annotations or contradictory annotations (e.g., a step marked as correct but assigned an error type) are also flagged for review and re-annotation. 
Besides, we conduct regular annotation review meetings where annotators discuss challenging cases and resolve ambiguities collaboratively.
We also develop a detailed annotation guideline document, which is continuously updated based on annotator feedback and edge cases encountered during the annotating process. 
The multi-layered approach ensures a high level of consistency and reliability in the final annotations.

\subsection{BreakDown Statistics}
\label{appendix_breakdown_statistics_ProJudgeBench}
\begin{table*}[tp]
  \centering
  \resizebox{0.78\textwidth}{!}{
  \begin{tabular}{lccccccccc}
    \toprule[0.8pt]
    \multirow{2}{*}{}
    & \multirow{2}{*}{\textbf{Overall}}  
    & \multicolumn{2}{c}{\textbf{Math}}  
    & \multicolumn{2}{c}{\textbf{Physics}}
    & \multicolumn{2}{c}{\textbf{Chemistry}}
    & \multicolumn{2}{c}{\textbf{Biology}}\\
    \cmidrule(){3-4} \cmidrule(){5-6} \cmidrule(){7-8} \cmidrule(){9-10} 
    && \textbf{K12} & \textbf{Comp} & \textbf{K12} & \textbf{Comp} & \textbf{K12} & \textbf{Comp} &\textbf{K12} & \textbf{Comp}   \\
     \midrule
    
    \# Samples & 2400 & 450 & 150 & 300 & 300 & 300 & 300 & 300 & 300 \\
    Avg. Steps & 20.8 & 23.7 & 21.3 & 21.5 & 26.0 & 20.0 & 19.9 & 15.2 & 18.0 \\
    Max. Steps & 470 & 470 & 215 & 199 & 353 & 209 & 197 & 78 & 150 \\
    \% Error Steps & 21.6 & 23.0 & 19.7 & 21.2 & 23.1 & 18.9 & 24.4 & 15.7 & 23.4 \\
    Avg. Error Steps & 6.6 & 8.1 & 7.0 & 7.0 & 8.4 & 5.7 & 6.2 & 4.2 & 5.7 \\
    Max. Error Types & 5 & 4 & 4 & 4 & 4 & 4 & 5 & 3 & 4 \\
    
    \bottomrule[0.8pt]
    
  \end{tabular}
  }
  \caption{Statistics of ProJudgeBench. K12 and Comp represent normal and competition-level problems, respectively.}
  \label{tab:statistics_ProJudgeBench}
\end{table*}
The breakdown statistics of ProJudgeBench is shown in Table~\ref{tab:statistics_ProJudgeBench}. 
We can see that, as the difficulty of the problems increases, the average number of steps per solution also rises. Besides, higher-difficulty problems exhibit a greater proportion of erroneous steps and more diverse error types, with competition-level math problems reaching up to 512 steps, indicating that models tend to generate more steps when tackling complex problems.
Higher-difficulty problems also exhibit a greater proportion of erroneous steps and more diverse error types, with some solutions containing up to 8 erroneous steps and 4 distinct error types.
This highlights the importance of fine-grained process evaluation: if the judge model can diagnose all these errors, it will enable a more comprehensive analysis of the model's weaknesses, offering targeted feedback for improvement.

\begin{figure}[!t]
\centering
  \resizebox{0.5\linewidth}{!} {
    \includegraphics{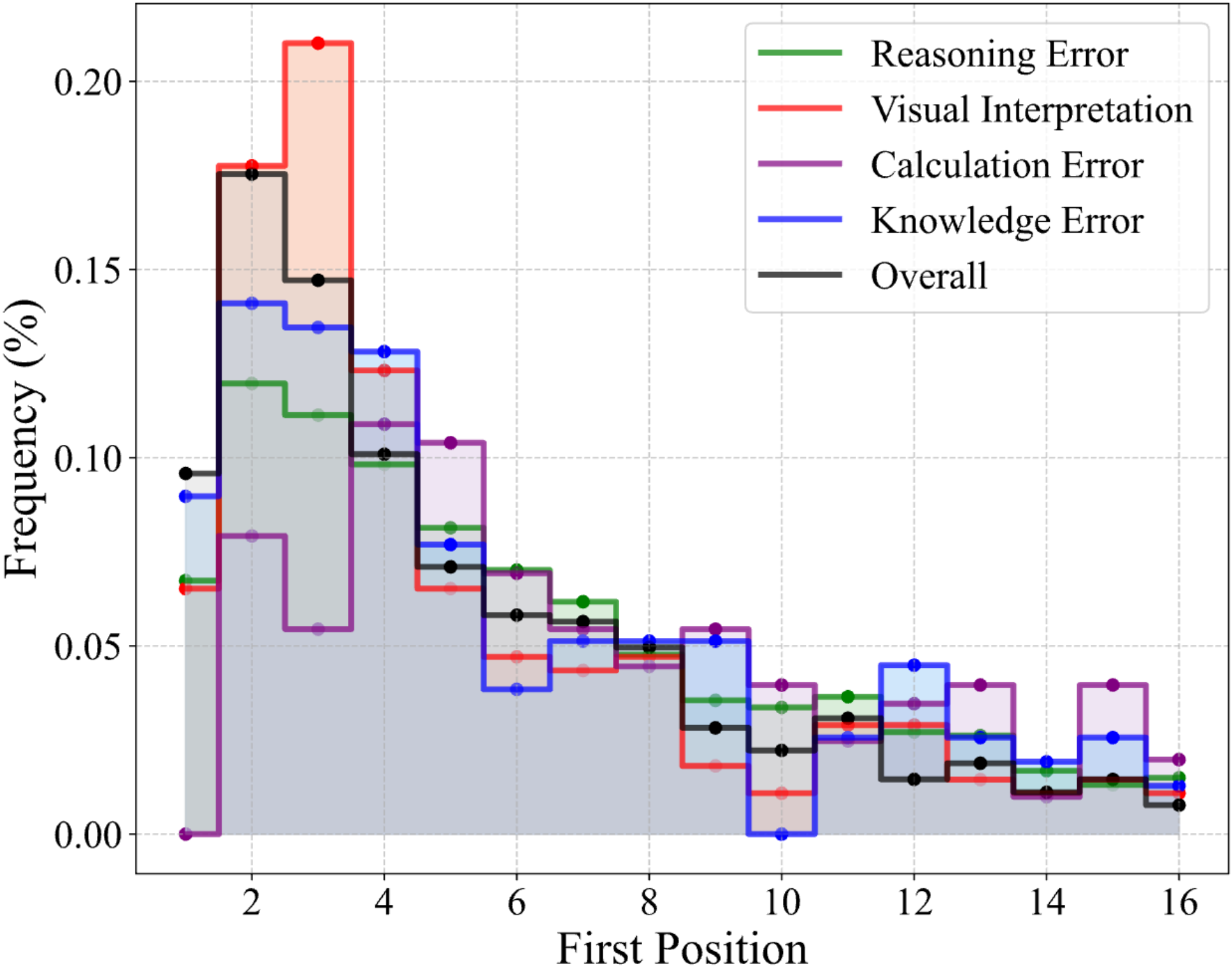}
  }
  \caption{
  Distribution of the first occurance steps for different error types. Truncated to 16 for better visualization.}
    \label{fig:error_types_position}
\end{figure}
We also plot the frequency of the first occurrence step for different error types in Figure~\ref{fig:error_types_position} . While the distribution varies across error categories, a consistent overall pattern emerges: errors peak in frequency during the earlier steps (e.g., steps 0-3) and gradually decline thereafter. Visual interpretation errors show a distinct peak at step 3, highlighting the model's initial struggles with accurately interpreting visual elements. Calculation errors, on the other hand, are more evenly distributed across all steps, indicating persistent challenges in performing precise mathematical operations throughout the reasoning process. In contrast, reasoning errors become more frequent in later steps, suggesting that models face increasing difficulty in maintaining logical consistency as the solution progresses.

\section{Detailed Information of ProJudge-173k}
In this section, we provide detailed information on ProJudge-173k, including the list of MLLMs used for generating solutions in data construction, data filtering process, and breakdown statistics.

\subsection{Data Construction}
\label{appendix_data_construction_ProJudge-173k}
As described in Section~\ref{sec:ProJudge-173k_data_construction}, we utilize 9 distinct MLLMs to generate solutions in ProJudge-173k. The list of utilized MLLMs are presented in Table~\ref{tab:MLLMs_list_ProJudge-173k}.
\begin{table*}[h!]
\centering
  \resizebox{0.6\textwidth}{!}{
    \begin{tabular}{l|l|l}
    \toprule[0.5pt]
\multicolumn{3}{c}{\textbf{MLLMs as solution generators in ProJudge-173k}} \\
\midrule
\tabincell{l}{
InternVL2.5-8B \\
InternVL2.5-26B \\
InternVL2.5-38B
} &
\tabincell{l}{
Qwen2.5-VL-Instruct-3B \\
Qwen2.5-VL-Instruct-7B \\
Qwen2.5-VL-Instruct-72B 
} &
\tabincell{l}{
MiniCPM-V-2\_6 (8B) \\
QVQ-72B-Preview \\
LLaVA-OneVision (7B) \\
}
\\

    \bottomrule[0.5pt]
    \end{tabular}
  }
    \caption{List of MLLMs used in ProJudge-173k to generate solutions.
  }
  \label{tab:MLLMs_list_ProJudge-173k}
\end{table*}

\subsection{Data Filtering}
\label{appendix_data_filtering_ProJudge-173k}
As described in Section~\ref{sec:ProJudge-173k_data_construction}, we conduct rigorous filtering processes to ensure data quality. 
Specifically, we apply three quality control mechanisms:
(1) \textbf{Format Consistency}: 
We remove samples that deviate from the predefined format, where each step in the student solution must be annotated as a tuple containing the step description, correctness, error type, and a brief explanation.
Additionally, samples with mismatched step counts between annotations and student solutions are discarded.
(2) \textbf{Annotation Consistency}: Samples with contradictory or incomplete annotations, such as steps marked as incorrect but lacking error type or cause descriptions, are excluded.
(3) \textbf{Error Coverage}: Samples with insufficient error diversity or repetitive error patterns are excluded to maintain dataset variety.

\begin{table}
  \centering
  \resizebox{0.5\textwidth}{!}{
  \begin{tabular}{@{}lcccc@{}}
    \toprule
             & Total & Camel-AI & K12 & Olympiad\\
    \midrule
    \# Samples & 173,354 & 26,249 & 93,184 & 53,921\\
    \# Problems & 26,084 & 8,000 & 12,000 & 6,084\\
    \# Math & 58,922 & 7,196 & 15,908 & 35,818\\
    \# Physics & 40,910 & 6,268 & 16,539 & 18,103\\
    \# Chemistry & 37,789 & 6,656 & 31,133 & 0\\
    \# Biology & 35,733 & 6,129 & 29,604 & 0\\
    Avg. Steps & 18 & 11 & 17 & 22 \\
    Max. Steps & 926 & 77 & 487 & 926 \\
    \% Error & 24.72 & 20. & 23.55 & 27.52\\
    \bottomrule
  \end{tabular}
  }
  \caption{Statistics of data sources and quantities in ProJudge-173k.}
  \label{tab:statistics_ProJudge-173k}
\end{table}
\subsection{BreakDown Statistics}
\label{appendix_breakdown_statistics_ProJudge-173k}
The statistics of the data sources and quantities are presented in Table~\ref{tab:statistics_ProJudge-173k}. 
By providing detailed error analysis, realistic reasoning paths, and diverse problem types, our dataset lays a solid foundation for advancing research in process evaluation, particularly for improving model's capabilities of error diagnosis in complex reasoning tasks.

\section{Detailed Information of Process Evaluation}

\subsection{Fine-tuning Details}
\label{appendix_fine_tuning_details}
In the experiments, we fine-tune InternVL2.5-8B, Qwen2.5-VL-3B-instruct and Qwen2.5-VL-7B-Instruct on ProJudge-173k with Dynamic Dual-Phase strategy. 

The training is conducted on 8 H100 GPUS. All the models are fine-tuned using LoRA for one epoch, with the same training set. The global batch size is set to 16, with per-device batch size of 4 and gradient accumulation steps of 2.
For InternVL2.5-8B, we employ a learning rate of 4e-5, optimized using a cosine learning rate scheduler with a warmup ratio of 0.03. Additionally, we applied weight decay of 0.05 to regularize the training process and prevent overfitting.
For Qwen2.5-VL-3B and Qwen2.5-VL-7B, the models are trained with a learning rate of 1.0e-4, also using a cosine learning rate scheduler and a warmup ratio of 0.1. 
Both fine-tuning processes utilize mixed-precision training (bf16) to accelerate computation and reduce memory usage. For InternVL2.5-8B, we additionally enable gradient checkpointing to further optimize memory usage during training.


\section{Prompts}
\subsection{Prompts for Injection Errors}
\label{appendix_prompts_inject_errors}
As discussed in Section~\ref{sec:ProJudge-173k_data_construction}, we utilize GPT-4o to intentionally inject errors into correct solutions. The prompts we use are displayed in Table~\ref{tab:prompt_inject_errors}.
\begin{longtable}
{p{\textwidth}}
\toprule
\subsubsection*{1. System Prompt} 
You are a highly experienced educator with a strong understanding of both solving problems and mimicking realistic, common mistakes made by students or AI.

\subsubsection*{2. User Content}
\textbf{Task:} Analyse the following question, tested knowledge points and reference step-by-step solution. \\
Introduce REASONABLE and REALISTIC errors into solution that mimic COMMON mistakes made by students or AI. \\
\\
\textbf{Instructions:} \\
1. Understand the following error categories and incorporate one or more of the following error types: \\
a. Numerical Calculation Error. Errors in basic arithmetic operations such as addition, subtraction, division, or square roots.\\
b. Symbolic Calculation Error. Errors in manipulating algebraic expressions, such as incorrect expansion, factoring, simplification, or solving equations with variables. \\
c. Visual Interpretation Error. Errors in interpreting graphical data, such as misidentifying coordinates, shapes, spatial relationships, or data within figures. \\
d. Reasoning Error. Errors in the logical thinking process that lead to incorrect conclusions, such as flawed arguments, invalid inferences, or gaps in the logical flow of the solution. \\
e. Knowledge Error. Errors caused by insufficient understanding or incorrect application of necessary knowledge (e.g., concepts, formulas, theorems, methods), or using outdated or incorrect information. \\
f. Question Understanding Error. Errors due to misunderstanding or misinterpreting the problems' conditions or requirements, such as misreading questions or misapplying given conditions. \\
\\
2. Introduce Errors: \\
a. Insert multiple errors into the reference solution steps. \\
b. Errors should be RESONABLE, REALISTIC and resemble COMMON mistakes, NOT arbitrary or overly obvious. \\
c. AVOID REPEATING the same error reason across steps. ENSURE each step is evaluated INDIVIDUALLT. \\
\\
3. Generate Erroneous Solutions: \\
a. Provide 3–5 erroneous solutions to cover diverse possibilities. \\
b. Ensure the erroneous solutions align with the question, remain logically consistent and misleading enough to challenge the reader. \\
\\
4. Response Format: Present each erroneous solution step-by-step in the following format. \\
a. Mark each step as 1 (Correct) or 0 (Incorrect). \\
b. For incorrect steps, specify the error type and a brief error description. \\
c. Example: \\
 $[$ \\
$ $ $ $ $ $ $ $ $ $ $ $ $ $ $ $          $[$ \\      
$ $ $ $ $ $ $ $ $ $ $ $ $ $ $ $ $ $ $ $ $ $ $ $ $ $ $ $ $ $ $ $            $[$"Step Description 1", 0, "Error category","a brief error descrition"$]$,   \\
$ $ $ $ $ $ $ $ $ $ $ $ $ $ $ $ $ $ $ $ $ $ $ $ $ $ $ $ $ $ $ $             $[$"Step Description 2", 1, "",""$]$,    \\ 
$ $ $ $ $ $ $ $ $ $ $ $ $ $ $ $          $]$, \\
 $]$\\
d. Strictly adhere to the format! DO NOT add any explanations, extra content, or annotations outside the specified format! \\
\\
\# Problem: \{problem\} \\
\\
\# Tested Knowledge points: \{knowledge points\} \\
\\
\# Reference solution: \{step-by-step reference solution\} \\
\\
\bottomrule
\caption{Prompt for injecting errors.}
\label{tab:prompt_inject_errors}
\end{longtable}

\subsection{Prompts for Solution Generation}
\label{appendix_prompts_solution_generation}
As described in Section~\ref{sec:ProJudgeBench_data_construction} and Section~\ref{sec:ProJudge-173k_data_construction}, we use diverse MLLMs to generate solutions, capturing a wide range of realistic reasoning behaviors and error patterns. The prompts we use for generating solutions are displayed in Table~\ref{tab:prompt_generate_solutions}.
\begin{longtable}
{p{\textwidth}}
\toprule
\subsubsection*{1. System Prompt} 
You are a highly skilled student proficient in solving scientific problems.
\subsubsection*{2. User Content}
Based on the given images, solve the following question. \\
\\
Here is some context information for this question, which might assist you in solving it: \{context\}* \\
\\
Problem: \{problem\}  \\
\\
Think step by step logically, considering all relevant information before answering.    \\
Write out the solution process, and use the same LaTeX format as the question in the solution process. \\
Please end your response with: "The final answer is \(\boxed{ANSWER}\)." \\
\\
\bottomrule
\caption{Prompt for generating solutions.}
\label{tab:prompt_generate_solutions}
\end{longtable}

.

\subsection{Prompts for Spliting Solutions into Steps}
\label{appendix_prompts_split}
To standardize the step granularity of each solution, we prompt Qwen2.5-72B-Instruct to split solutions into logically complete and progressive steps. The prompts we use for splitting are displayed in Table~\ref{tab:prompt_split}.
\begin{longtable}
{p{\textwidth}}
\toprule
\subsubsection*{User Content:}
Please split the following solution steps into individual steps and return them formatted as a Python list. \\
Each step should be a separate string within the list. \\
If the solution steps contain only a single step or sentence, DO NOT split it further, return it as a single element in the list. \\
Only return the list itself, with no additional text or formatting. \\
DO NOT modify the text in any way, simply split it. \\
\\
Example Format: \\
$[$ \\
$ $ $ $ $ $ $ $ $ $ $ $ $ $ $ $"First step description", \\
$ $ $ $ $ $ $ $ $ $ $ $ $ $ $ $"Second step description", \\
$ $ $ $ $ $ $ $ $ $ $ $ $ $ $ $"Third step description", \\
$ $ $ $ $ $ $ $ $ $ $ $ $ $ $ $... \\
$]$ \\
\\
Solution steps to split: {solution} \\
\\
\bottomrule
\caption{Prompt for generating solutions.} \\
\label{tab:prompt_split}
\end{longtable}

\subsection{Prompts for Process Evaluation}
\label{appendix_prompts_process_evaluation}
As described in Section~\ref{sec:experiments_setup}, we use the same evaluation for all models to ensure consistency. The prompts for process evaluation are displayed in Table~\ref{tab:prompt_process_evaluation}.
\begin{longtable}
{p{\textwidth}}
\toprule
\subsubsection*{1. System Prompt} 
You are a teacher skilled in evaluating the intermediate steps of a student’s solution to a given problem.

\subsubsection*{2. User Content}
You are given a scientific problem, its correct final answer, and a student's solution to evaluate. \\
Your task is to: first, solve the problem yourself, using the correct final answer as a hint. Ensure your reasoning leads to the correct answer.
Once you have a clear understanding of how the problem could be solved, evaluate the correctness of each step in the student’s solution. \\
Focus exclusively on the scientific, logical, or mathematical correctness of the solution. Ignore differences in formatting, expression style, specific wording, or presentation order, as long as the reasoning and results are valid. \\
\\
For each step, perform: \\
1. Binary scoring: assign a score of 1 for correct steps and 0 for incorrect steps. \\
2. Error classification (only if the step is incorrect): \\
    a. Numerical Calculation Error. Errors in basic arithmetic operations such as addition, subtraction, division, or square roots.\\
b. Symbolic Calculation Error. Errors in manipulating algebraic expressions, such as incorrect expansion, factoring, simplification, or solving equations with variables. \\
c. Visual Interpretation Error. Errors in interpreting graphical data, such as misidentifying coordinates, shapes, spatial relationships, or data within figures. \\
d. Reasoning Error. Errors in the logical thinking process that lead to incorrect conclusions, such as flawed arguments, invalid inferences, or gaps in the logical flow of the solution. \\
e. Knowledge Error. Errors caused by insufficient understanding or incorrect application of necessary knowledge (e.g., concepts, formulas, theorems, methods), or using outdated or incorrect information. \\
f. Question Understanding Error. Errors due to misunderstanding or misinterpreting the problems' conditions or requirements, such as misreading questions or misapplying given conditions. \\
g. No solution provided. The model refuses to answer, fails to follow instructions to make a solution, or encounters anomalies in generation process such as repetitive responses or incomplete outputs. \\
3. Provide a brief explanation for the identified error. \\
\\
\# The given problem: \{problem\} \\
\\
\# The Correct Final Answer: \{final answer\} \\
\\
\# Student's solution: {step-by-step student's solution} \\
\\
Finally, your evaluation results should be a Python list within $<$evaluation$>$ and $</$evaluation$>$ tags, as follows: \\
$<$evaluation$>$ \\
$[$ \\
$ $ $ $ $ $ $ $ $ $ $ $ $ $ $ $    $[$"The full text of a correct step", 1, "", ""$]$, \\
$ $ $ $ $ $ $ $ $ $ $ $ $ $ $ $     $[$"The full text of an incorrect step", 0, "Error category", "Brief error descrition"$]$, \\
$]$ \\
$</$evaluation$>$ \\
Strictly adhere to the output format, with no additional text or formatting. \\
Ensure the length of your output list matches the student's solution. \\
\\
\bottomrule
\caption{Prompt for process evaluation.}
\label{tab:prompt_process_evaluation}
\end{longtable}

\end{document}